\documentclass[lettersize,journal]{IEEEtran}
\usepackage{amsmath,amsfonts}
\usepackage{algorithmic}
\usepackage{array}
\usepackage[caption=false,font=normalsize,labelfont=sf,textfont=sf]{subfig}
\usepackage{textcomp}
\usepackage{url}
\usepackage{verbatim}
\usepackage{graphicx}

\usepackage{ifpdf}

\usepackage{cite}
\usepackage{bbm}

\usepackage{lastpage}
\usepackage{fancyhdr}

\usepackage{hyperref}
\hypersetup{hidelinks,
	colorlinks=false,
	allcolors=black,
	pdfstartview=Fit,
	breaklinks=true}

\usepackage[switch]{lineno}

\usepackage{amsmath,amssymb,amsfonts}
\usepackage{float}
\usepackage{algorithmic}

\usepackage{array}

\usepackage{textcomp}
\usepackage{xcolor}
\usepackage{tabularray}

\usepackage{color}

\usepackage{dblfloatfix}
\usepackage{url}

\usepackage[utf8]{inputenc}
\usepackage{graphicx}
\usepackage{caption}

\usepackage{float}

\usepackage{multicol}
\usepackage{multirow}
\usepackage{threeparttable}

\usepackage{color}
\usepackage{multirow}
\usepackage{dsfont}

\PassOptionsToPackage{prologue,dvipsnames}{xcolor}
\usepackage{booktabs}
\usepackage{colortbl}  %彩色表格需要加载的宏包
\usepackage{xcolor}
\usepackage{array}   %对表列和表格线的设置需要用到array宏包

\usepackage{fancyhdr} % 导入 fancyhdr 宏包
\usepackage[absolute,overlay]{textpos} % 用于绝对定位文字

\usepackage{multirow}
\usepackage{colortbl} 
\usepackage{threeparttable}
\usepackage{threeparttablex}
\usepackage{algorithmic}
\usepackage{algorithm}

\newcommand{\PYCOMMENT}[1]{\textcolor[HTML]{8fbcc2}{\# #1}}

\def\BibTeX{{\rm B\kern-.05em{\sc i\kern-.025em b}\kern-.08em
    T\kern-.1667em\lower.7ex\hbox{E}\kern-.125emX}}
\usepackage{balance}
\begin{document}

\title{Adaptive Spatial Augmentation for Semi-supervised Semantic Segmentation}

\author{Lingyan Ran, Yali Li, Tao Zhuo, Shizhou Zhang, Yanning Zhang,~\IEEEmembership{Fellow,~IEEE}

\thanks{This work is supported in part by the National Natural Science Foundation of China (62476226)
% , Natural Science Basic Research Program of Shaanxi (2024JC-YBQN-0719), Natural Science Foundation of NingBo (2023J262)
. \textit{(Corresponding author: Tao Zhuo.)}}
\thanks{Lingyan Ran, Yali Li, Shizhou Zhang, and Yanning Zhang are with the Shaanxi Provincial Key Laboratory of Speech and Image Information Processing, and the National Engineering Laboratory for Integrated Aerospace-Ground-Ocean Big Data Application Technology, School of Computer Science, Northwestern Polytechnical University, Xi’an 710072, China.
Tao Zhuo is in the College of Information Engineering, Northwest A\&F University, Yangling, 712100, China.
}
}

\markboth{Journal of \LaTeX\ Class Files,~Vol.~18, No.~9, September~2020}%
{How to Use the IEEEtran \LaTeX \ Templates}

\maketitle

\begin{abstract}
In semi-supervised semantic segmentation (SSSS), data augmentation plays a crucial role in the weak-to-strong consistency regularization framework, as it enhances diversity and improves model generalization. Recent strong augmentation methods have primarily focused on intensity-based perturbations, which have minimal impact on the semantic masks. In contrast, spatial augmentations like translation and rotation have long been acknowledged for their effectiveness in supervised semantic segmentation tasks, but they are often ignored in SSSS. In this work, we demonstrate that spatial augmentation can also contribute to model training in SSSS, despite generating inconsistent masks between the weak and strong augmentations. Furthermore, recognizing the variability among images, we propose an adaptive augmentation strategy that dynamically adjusts the augmentation for each instance based on entropy. Extensive experiments show that our proposed Adaptive Spatial Augmentation (\textbf{ASAug}) can be integrated as a pluggable module, consistently improving the performance of existing methods and achieving state-of-the-art results on benchmark datasets such as PASCAL VOC 2012, Cityscapes, and COCO.
\end{abstract}


\section{Introduction}
\IEEEPARstart{S}{emantic} segmentation focuses on assigning semantic labels to every pixel in an image, and it has been widely used in various domains, including analyses of natural images~\cite{ma2025dual,zhang2025frequency}, medical images~\cite{ding2024clustering}, remote sensing~\cite{lran2024DDF}, 
autonomous driving~\cite{guo2024vanishing}, etc.
% , and point cloud data~\cite{cheng2021sspc,li2024construct}.
The success of supervised semantic segmentation methods depends on a large collection of high-quality pixel-level annotated images for training. 
However, the process of pixel-wise labeling is both labor-intensive and time-consuming. 
This challenge has led to the rise of semi-supervised learning (SSL)~\cite{chen2025towards,lu2025uncertainty} in semantic segmentation, which utilizes both annotated and unannotated data to train models, providing a more efficient and scalable approach. 

% Incorporating deep learning has notably enhanced the efficiency of these tasks; however, 
% it necessitates substantial labeled datasets for model training. Despite the availability of vast amounts of raw data, the manual labeling process, entailing pixel-by-pixel annotation, is labor-intensive and time-consuming. Consequently, training models for applications in open-world scenarios with limited annotations presents a significant challenge. 
% Semi-supervised learning (SSL)~\cite{chen2025towards,lu2025uncertainty}, which leverages labeled and unlabeled data, offers a compelling solution.
% % tarvainen2017mean
% This approach has led to the creation of numerous novel techniques in semi-supervised semantic segmentation (SSSS).

\begin{figure}[!t]
    \centering
    \includegraphics[width=1\linewidth]{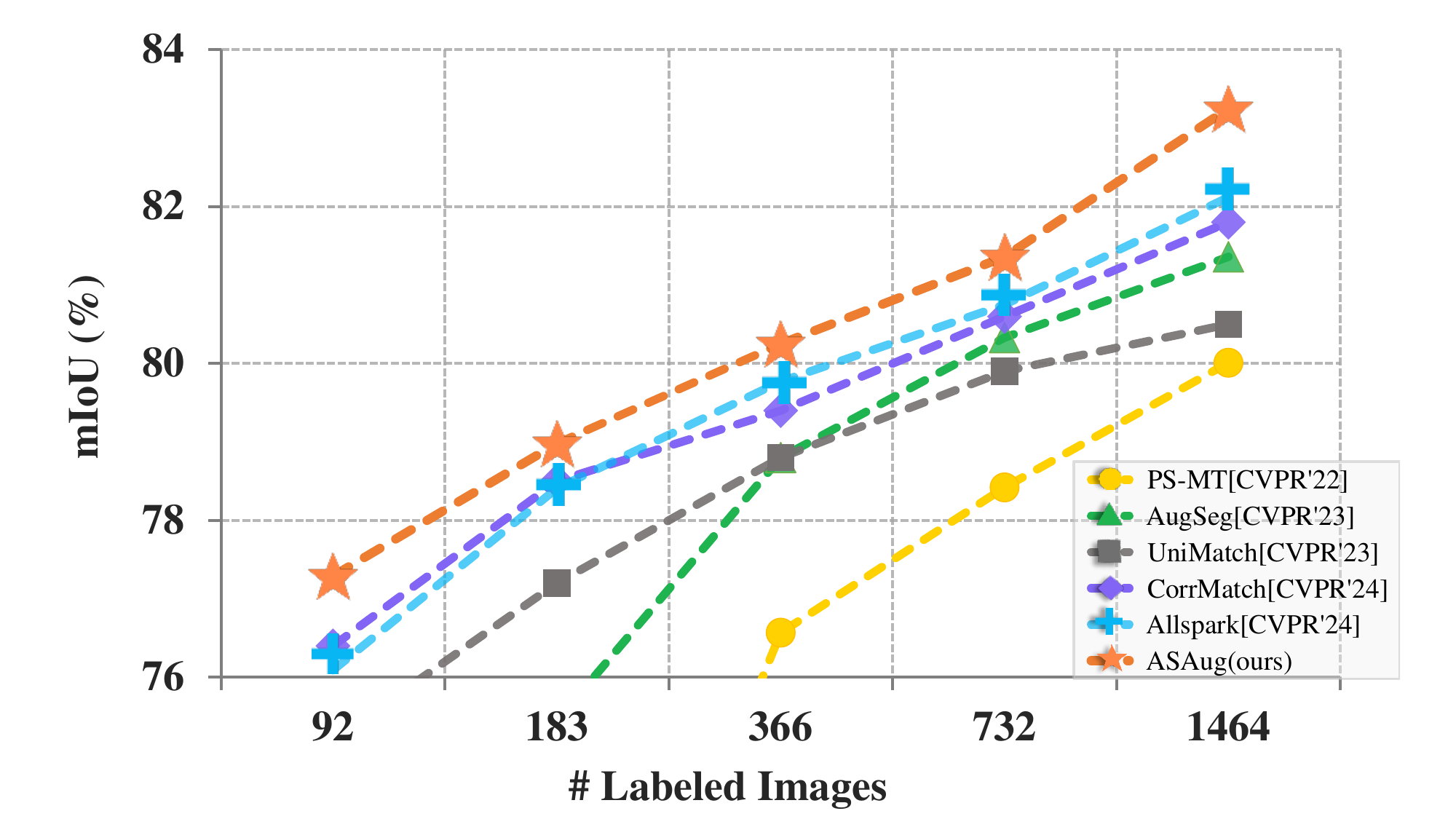}
    \vspace{-1.5em}
    \caption{Comparison with SOTA methods on the Pascal VOC 2012 dataset. Notably, our method outperforms other approaches in all partitioning scenarios.
    % (ASAug results are implemented on the Allspark\cite{wang2024allspark}).
    }
    \label{fig_trend}
    \vspace{-1.0em}
\end{figure}

% 过渡
Recent semi-supervised semantic segmentation (SSSS) approaches~\cite{pseudolabel,hu2024multi,xiao2024multi} often adopt the weak-to-strong consistency regularization framework~\cite{zou2021PseudoSeg,lee2021anti,yang2023revisiting}. These methods train a teacher model on weakly augmented data and a student model on strongly augmented data, ensuring consistent outputs despite the perturbations introduced by the augmentations. Weak augmentations typically include operations like image flipping, cropping, and scaling, while strong augmentations are generally intensity-based~\cite{yuan2021simple, zhao2023augmentation}, such as identity, auto-contrast, Gaussian blur, equalize, sharpness, brightness, hue, color jitter, posterize, and solarize. As highlighted in~\cite{zhao2023augmentation}, strong augmentations are a crucial component that enhances model training.

% Yuan et al.~\cite{yuan2021simple} demonstrate the critical role of strong augmentations within a straightforward SSL setup. AugSeg~\cite{zhao2023augmentation} designs a highly stochastic intensity-based augmentation method by randomly selecting data transformations and sampling distortion strengths uniformly from a continuous space. Ghiasi et al~\cite {ghiasi2021simple} employ copy-paste as a form of strong augmentation. 

% ！！！ novelty 再次总结！！！
This study finds that commonly employed spatial augmentations in supervised semantic segmentation, like rotation and translation~\cite{Cubuk_2019_CVPR,cubuk2020randaugment,muller2021trivialaugment}, are often not utilized for strong augmentation in semi-supervised scenarios. Spatial augmentations differ from intensity-based ones by altering pixel positions, thus changing the mask and potentially causing discrepancies between weak and strong augmentations. This observation prompts the following inquiry: can spatial augmentations enhance SSSS comparably to their impact on the supervised methods?

\begin{table}[]
\footnotesize
% \small
\centering
\caption{Comparison of recent SSSS solutions in terms of “More Supervision", “Augmentation", and “Hybrid Techniques" which are combined with other algorithms (in order of publication year). We explain the acronyms as follows. “\textbf{CATP}": co-training/auxiliary trainable-network/pseudo-rectifying, including dual-model co-training and trainable auxiliary networks, “\textbf{MTS}”: multiple training stages. “\textbf{SDA}": strong data augmentation, “\textbf{MIE}”: multi-stream/multi-branch interaction/evaluation. “\textbf{CR}": consistency regularisation, here mainly refers to introducing perturbations. “\textbf{CL}”: contrast learning. ASAug has a simple enough architecture and better performance.
}
\begin{tabular}{c|cc|cc|cc}
% \hline
\toprule[1pt]
% \rowcolor[HTML]{ECF4FF} 
% \rowcolor[HTML]{FFFFFF} 
\cellcolor[HTML]{ECF4FF} &
  \multicolumn{2}{c|}{\cellcolor[HTML]{ECF4FF}\textbf{More Sup.}} &
  \multicolumn{2}{c|}{\cellcolor[HTML]{ECF4FF}\textbf{Aug.}} &
  \multicolumn{2}{c}{\cellcolor[HTML]{ECF4FF}\textbf{Hybrid-tec.}} 
  \\ 
  \cline{2-7}
  % \cmidrule{2-7}
\rowcolor[HTML]{ECF4FF} 
\multirow{-2}{*}{\cellcolor[HTML]{ECF4FF}\textbf{Method}} &
  \textbf{CATP} &
  \textbf{MTS} &
  \textbf{SDA} &
  \textbf{MIE} &
  \textbf{CR} &
  \textbf{CL} \\
  \midrule[1pt]
  % \hline
CCT\cite{ouali2020semi}           & \checkmark &   &            &   & \checkmark &   \\ 
% \cline{1-1}
% ECS\cite{mendel2020semi}          & \checkmark &   &            &   &   &   \\ 
% \cline{1-1}
C3-semi.\cite{zhou2021c3}       &   &   & \checkmark          &   & \checkmark & \checkmark \\ 
% \cline{1-1}
DMT\cite{feng2022dmt}             & \checkmark &   & \checkmark          &   &   &   \\ 
% \cline{1-1}
RoCo\cite{liu2021bootstrapping}   &   &   & \checkmark          &   &   & \checkmark \\ 
% \cline{1-1}
CPS\cite{chen2021semi}            & \checkmark &   & \checkmark          &   &   &   \\ 
% \cline{1-1}
ST++\cite{yang2022st++}           &   & \checkmark & \checkmark          &   &   &   \\ 
% \cline{1-1}
ELN\cite{kwon2022semi}            & \checkmark & \checkmark & \checkmark          &   &   &   \\ 
% \cline{1-1}
PS-MT\cite{liu2022perturbed}      &   &   & \checkmark          &   & \checkmark &   \\ 
% \cline{1-1}
U2PL\cite{wang2022semi}           &   &   & \checkmark          &   &   & \checkmark \\ 
% \cline{1-1}
UCC\cite{fan2022ucc}              &   &   & \checkmark          & \checkmark & \checkmark &   \\ 
% \cline{1-1}
UniMatch\cite{yang2023revisiting} &   &   & \checkmark          & \checkmark & \checkmark  &   \\
% \cline{1-1}
CCVC\cite{wang2023conflict}       
& \checkmark &   &  & \checkmark & \checkmark &   \\ 
iMAS\cite{zhao2023instance}
& &   &\checkmark  & \checkmark &  &   \\
% \cline{1-1}
CISC-R\cite{wu2023querying}       &   & \checkmark & \checkmark          &   &   &   \\ 
CorrMatch\cite{sun2024corrmatch}     
& \checkmark  &  & \checkmark          &   &   &   \\
Allspark\cite{wang2024allspark}       &   &  & \checkmark          & \checkmark  &   &   \\
% \hline
\midrule[1pt]
\rowcolor[HTML]{ECF4FF} 
\textbf{ASAug}        &   &   & \textbf{\checkmark} &   &   &   \\ 
\bottomrule[1pt]
% \hline
\end{tabular}
\label{tab: comp}
\end{table}

To address this question, we first conducted experiments by replacing intensity-based augmentations with spatial-based ones, specifically rotation and translation (see Table \ref{ablation_component}). Unexpectedly, our findings reveal that the inconsistent masks produced by these spatial augmentations enhance generalization. This improvement likely stems from the segmentation model’s need to accurately identify target regions under varying conditions, as spatial augmentations create a distinct gap between weak and strong augmentation scenarios.
Furthermore, recognizing the significant variability among individual instances, we introduce an adaptive spatial augmentation (ASAug) technique that leverages entropy to modulate augmentation strength, thereby reinforcing consistency regularization. 
This approach draws inspiration from methods such as autoaugment~\cite{Cubuk_2019_CVPR}, randaugment~\cite{cubuk2020randaugment}, and trivialaugment~\cite{muller2021trivialaugment}, all of which aim to automatically discover more effective data augmentation policies for image classification.
Since our method is pluggable, it can be easily integrated into existing weak-strong consistency regularization frameworks. Extensive experiments on three benchmark datasets demonstrate the effectiveness of our process to improve performance and achieve SOTA performance compared to other methods.

% Second, in the image segmentation task, it is crucial to accurately categorize each pixel in the image into its corresponding category, and thus the segmentation model is particularly sensitive to the spatial structure of the image. In real-world scenarios, due to frequent changes in object pose and position, the segmentation model needs to have the ability to maintain accurate recognition of the target region under different geometric variation conditions. Geometric transformations (rotations and translations) help the model learn to adapt to these changes and thus have stronger generalization in various changing scenarios. Therefore, our research focuses on effectively using geometric enhancement to improve the robustness and generalization performance of models.
% Finally, based on the above analysis, we design a dynamic enhancement pipeline based on consistency training, which mainly applies geometric enhancement to unlabeled data to change its pixel location points. The strength of the enhancement is dynamically adjusted through instance-level evaluations to increase the diversity of the data and enhance the model's ability to cope with complex scenes.

As shown in Table~\ref{tab: comp}, unlike previous strategies, we break away from traditional enhancement techniques by focusing on the essence of the segmentation task and practically analyzing the importance and substantial effect of enhancement. 

% 我们的贡献
Our contributions can be summarized in three folds:
\begin{itemize}
    % \item We find that the spatial-based augmentation strategy can also benefit model training in SSSS, which is often ignored in existing methods.

    \item To the best of our knowledge, this is the first study to demonstrate that spatial augmentations, often overlooked in existing SSSS methods, can boost generalization, even though they produce inconsistent masks.
    
    \item We introduce a pluggable ASAug module that adaptively applies strong augmentations to each instance with entropy-based adaptive weight (EAW), improving consistency regularization.
    \item Extensive experiments on three well-known benchmarks, PASCAL VOC 2012 \cite{everingham2010pascal}, Cityscapes \cite{cordts2016cityscapes}, and COCO\cite{lin2014microsoft}, demonstrate that our method surpasses the state-of-the-art methods by a noticeable margin.
\end{itemize}

% 对比图
\begin{figure}[!t]
    % \vspace{-1.0em}
    \centering
    \includegraphics[width=1.0\linewidth]{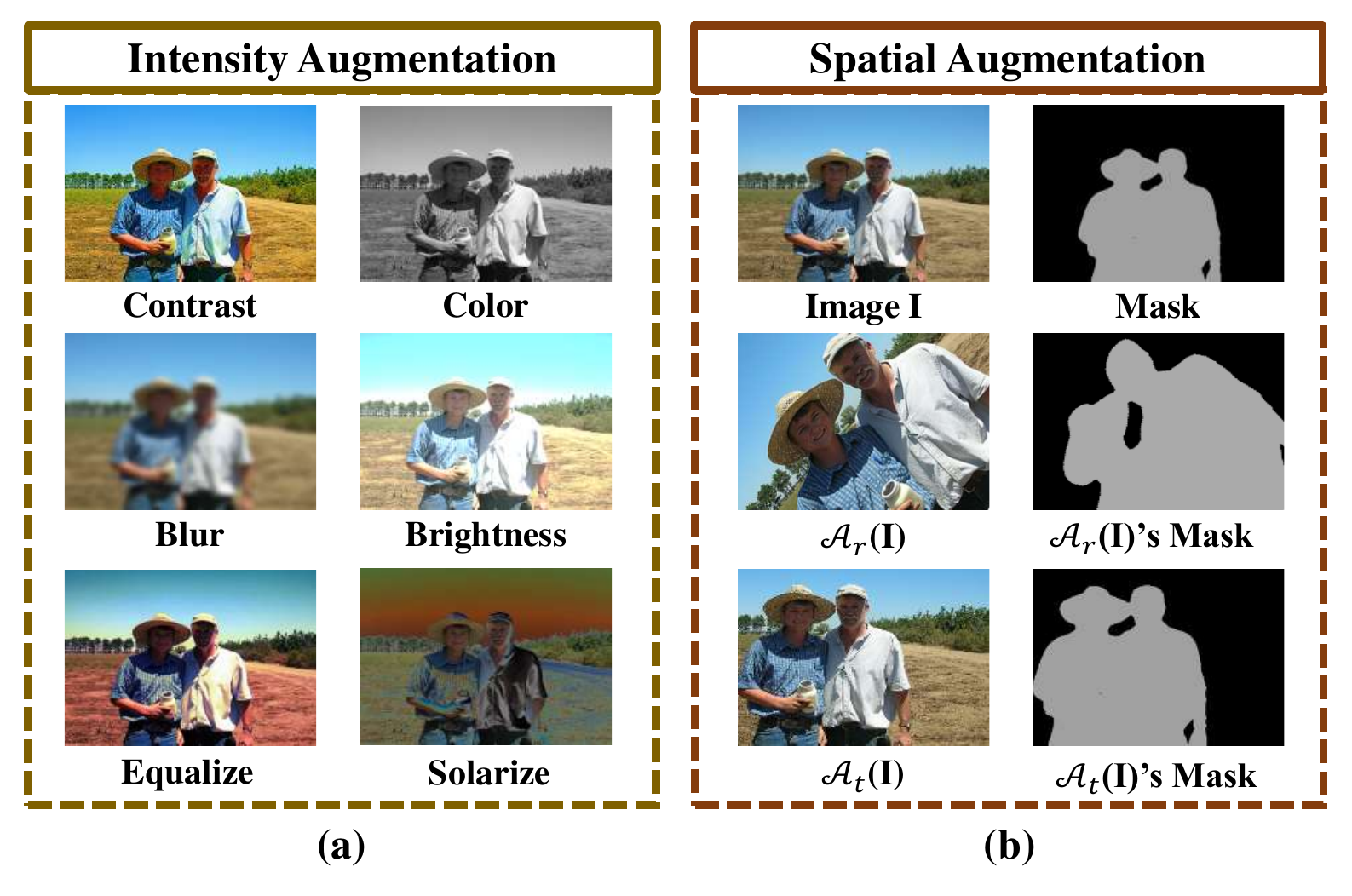}
    \vspace{-1.5em}
    \caption{Comparisons between the intensity and spatial augmentations. (a) intensity-based augmentations, like contrast, color, blur, brightness, and contrast adjustments, modify pixel appearance without changing spatial positions of the original mask; (b) spatial augmentations, such as rotation and translation, directly change the positions of pixels, leading to inconsistent masks between the original image and the augmented image.}
    \label{fig_comp}
    \vspace{-1.0em}
\end{figure}

\section{Related Work}

% 1 半监督学习
\subsection{Semi-Supervised Learning}
Semi-supervised learning(SSL) predominantly focuses on training models with a limited quantity of labeled data alongside a substantial volume of unlabeled data. Recent research branches into two principal categories: consistency regularization \cite{ouali2020semi,xu2021dash} and pseudo-labeling \cite{lee2013pseudo,adasemicd}. 
Consistency regularization entails training models to achieve high robustness by learning from both labeled and unlabeled data through perturbations, whereas pseudo-labeling aims to expand the dataset by having the model generate pseudo-labels for unlabeled data, subsequently utilizing both true and pseudo-labeled data for training. FixMatch \cite{sohn2020fixmatch} synthesizes these strategies by creating a hybrid method that leverages weakly augmented data for pseudo-label generation and employs cross-entropy for consistency regularization. Following this, FlexMatch \cite{NEURIPS2021_995693c1} introduced curriculum pseudo-labels, which modify category thresholds without adding extra parameters or computations.
% 
% 我们方法简述
To address the heavy reliance on labeled data, our proposed approach integrates SSL to significantly improve the performance of semantic segmentation models while reducing the need for annotated data.

% 2 半监督语义分割
\subsection{Semi-Supervised Semantic Segmentation}
SSL for semantic segmentation delivers outstanding performance and alleviates the challenge of limited labeled data for researchers. The survey~\cite{pelaez2023survey} introduces a taxonomy that systematically categorizes these techniques into pseudo-labeling (PL)~\cite{lee2013pseudo,pseudolabel}, consistency regularization (CR) \cite{bachman2014learning}, contrastive learning (CL) \cite{oord2018representation}, adversarial training (AT) \cite{goodfellow2014generative}, and hybrid methods (HM) \cite{hou2022semi}.

% 具体的分类
% GAN
Eary AT-related research utilized a common framework known as Generative Adversarial Networks (GANs)~\cite{xu2022self}, which can be chiefly categorized into two types depending on their structure: those with Generators\cite{souly2017semi,li2021semantic} and those lacking Generators\cite{cao2022adversarial,jin2021adversarial}.
% 对比学习
Subsequently, the scientists explored leveraging CL to gain more meaningful representations within the embedding space to enhance segmentation results. RoCo \cite{liu2021bootstrapping} addresses the ambiguity among confusion classes by employing contrast learning at the region level, while PRCL \cite{xie2023boosting} suggests interpreting contrast learning through probability expressions.
% to mitigate the risk of unreliable pseudo-labels through Gaussian modeling.
% 伪标签
PL enhances the labeled dataset by incorporating pseudo-labels and is gaining traction because of its straightforward implementation and interpretability. The study ST++ \cite{yang2022st++} introduces a reliability retraining method that involves multiple stages, while CISC-R \cite{wu2023querying} recommends querying labeled images to refine inaccurate pseudo-labels.
% 一致性正则
In the field of CR, there is a consensus on the smoothing assumption, which entails that identical predictions should be obtained for a pixel and its perturbed version. Building on FixMatch \cite{sohn2020fixmatch}, both PseudoSeg \cite{zoupseudoseg} and UniMatch \cite{yang2023revisiting} apply strong and weak consistency to segmentation tasks. Additionally, the integration of these techniques has resulted in notable outcomes. U2PL~\cite{wang2022semi} and DGCL \cite{wang2023hunting} merge PL and CL to optimize pseudo-labels via feature similarity between samples, thereby steering the model toward learning more precise and robust feature representations. 
A prevalent approach \cite{sohn2020fixmatch,wang2023conflict} nowadays involves integrating PL with CR under a consistency framework aimed at enhancing the confirmation bias and potentially boosting accuracy further.
% \cite{arazo2020pseudo} 
PC2Seg\cite{zhong2021pixel} uses feature space comparison learning and consistency training.

% 结构图
\begin{figure*}[!t]
    \vspace{-1.8em}
    \centering
    \includegraphics[width=0.85\linewidth]{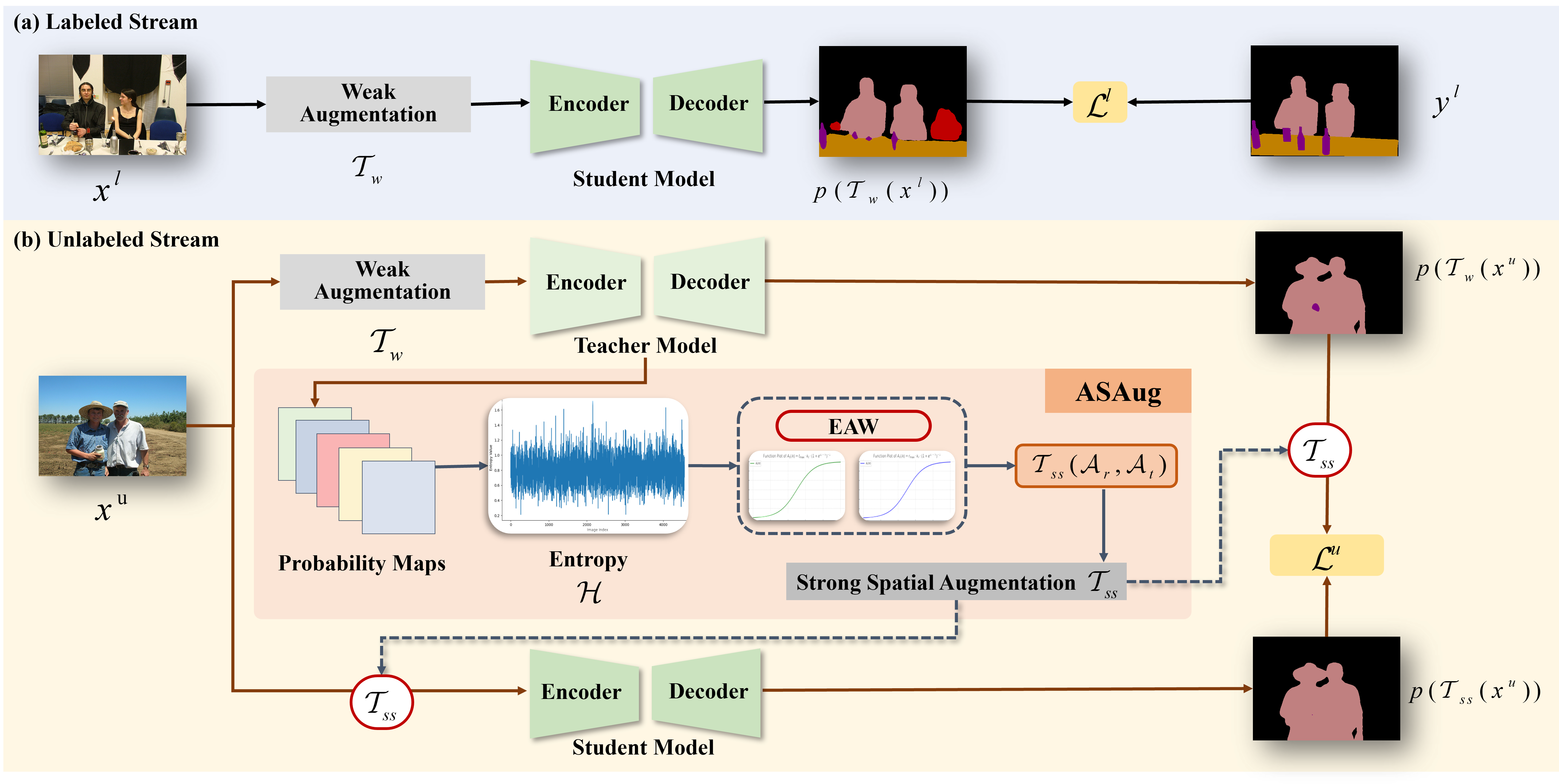}
    \vspace{-0.2em}
    \caption{Illustration of our ASAug pipeline. 
Based on the teacher-student consistency training framework \cite{ouali2020overview}, we introduce an adaptive method that can selectively distort images as geometrically strong enhancements based on their reliability and importance. ``EAW" denotes our entropy-based adaptive weight. Notably, geometric adjustment changes pixel point locations, and we apply the same operation to the teacher's output to ensure the consistency of predictions before and after enhancement.}
    \label{fig_struct}
    \vspace{-1.0em}
\end{figure*}

% 修改图
% 太素了
% 图里增强符号更新
% 两阶段分开

% 细节给出来 ASAug
% 画一个小图 展示 EAW
% 想想怎么改图

% 放 arxiv 上

% SSSS中的数据增强
\subsection{Data Augmentation Study in SSSS} 
\vspace{-0.5em}
Sufficient data is essential for tasks using deep learning methods, such as image classification, semantic segmentation, and change detection. With limited actuation, data augmentation~\cite{Cubuk_2019_CVPR,yuan2021simple} is widely used in various scenarios. 
Methods like random shuffle or intensity change are the most common. 
Although many approaches have been developed in SSSS, ST++ \cite{yang2022st++} demonstrates that the simplest self-training pipeline in PL coupled with strong data augmentation, can achieve excellent performance without any modular improvement. This proves that data augmentation plays a crucial role with limited labeled data.
% 传统增强工作
Yuan et al.~\cite{yuan2021simple} introduced a simple yet efficient framework as a baseline for robust data enhancement techniques. CutMix \cite{yun2019cutmix} effectively exploited the regularization effect of training pixels and the loss of retained regions by cutting and pasting patches in the training image, and subsequent ClassMix \cite{olsson2021classmix} and ComplexMix \cite{chen2021complexmix} were improved on this basis.
% AugSeg\cite{zhao2023augmentation} is guided by the confidence level, and various data enhancement techniques are randomly selected to adaptively enhance the samples. 
AugSeg~\cite{zhao2023augmentation} adaptively augmented samples based on confidence by randomly selecting various techniques.
They also argue that most research ignores the variability between instances and proposes an instance-specific model adaptive supervision method called iMAS \cite{zhao2023instance}.

% 几何增强工作 
As shown in Fig.~\ref{fig_comp}, traditional augmentation aims to artificially increase the diversity of the training data by performing various transformations or perturbations on the existing dataset. However, it may not adequately reflect the complexity and diversity required for a robust semantic segmentation model. 
Several researchers have noted the effect of positional factors on segmentation tasks and proposed geometric enhancements.
Cao et al. \cite{cao2022adversarial}proposed a context-aware unsupervised strategy for micro-aggregate warping. 
M3L \cite{grubivsic2023revisiting} introduces a robust perturbation model incorporating geometric warping and photometric variations.
MR-PhTPS \cite{grubivsic2021baseline} relies on nonlinear geometric and photometric perturbations.
% 我们方法简述
Unlike the above methods, we address the issue of the importance of geometric or positional factors on SSSS and accordingly propose an insertable adaptive mapping augmentation module that improves the efficiency of using unlabeled data through geometrically strong augmentation, which has been neglected in previous work.

\section{Method}
\label{sec:formatting}
The proposed method is built upon the weak-to-strong consistency regularization framework~\cite{yang2023revisiting,zhao2023instance,wang2024allspark,sun2024corrmatch}. Unlike recent approaches that primarily focus on intensity-based augmentation, we propose that spatial-based augmentations can also enhance model generalization. Fourthmore, taking into account the variability among samples, we introduce an adaptive augmentation strategy. The details of our approach are introduced below.

% 伪代码引入
% In addition, to explain our ASAug in more detail, we show the relevant pseudo-code in the algorithm ~\ref{alg_pseudo} in a PyTorch-like style to facilitate the reader's understanding.

In addition, to explain our ASAug in more detail, we show the overall architecture of our approach in Fig.~\ref{fig_struct}, and the associated pseudo-code in Algorithm~\ref{alg_pseudo} in a PyTorch-like style for ease of understanding.

% 伪代码
\begin{algorithm}[!t]
% \vspace{-1.0em}
\ttfamily
% \footnotesize
\small
% \normalsize
\caption{Pseudocode of ASAug in a PyTorch-like Style}
\begin{algorithmic}[0] % [0] 禁用行号
\STATE \PYCOMMENT{$f / f_t$: model / teacher\_model}\\
% , composed of an encoder $g$ and decoder $h$} \\
\STATE \PYCOMMENT{$aug\_w$: weak enhancements}
\STATE \PYCOMMENT{$aug\_spatial$:spatial enhancement ($\mathcal{A}_{r}$\&$\mathcal{A}_{t}$)}
\STATE \PYCOMMENT{$k_r / k_t$,$d_r / d_t$: scaling, offset parameters}
\STATE \PYCOMMENT{$r\_max / t\_max$: maximum rotation angle and maximum translation ratio}
\STATE \PYCOMMENT{$B$: batch size, $C$: number of categories, $H / W$: height/width of image}

\vspace{1em}

\FOR{each $x \in loader\_u$:}
    % \STATE \PYCOMMENT{predictions from the teacher model} \\
    \STATE x\_w = $aug\_w$(x) \PYCOMMENT{$[B, H, W]$}
    \STATE x\_w\_probmap = $f_t$(x\_w) 
    \PYCOMMENT{$[B, C, H, W]$} \\
    
    \STATE \PYCOMMENT{calculate the entropy}
    % \STATE \PYCOMMENT{calculate an entropy to determine the strength of the spatial enhancement}
    % \STATE x\_w\_probmap = $f_t$($aug\_w$)  
    % \PYCOMMENT{$[B, C, H, W]$} \\
    \STATE p\_w = x\_w\_probmap.argmax(dim=1)
    \STATE entropy = $comp\_entropy$(x\_w\_probmap) \\
    % \PYCOMMENT{$[B, H, W]$} \\
    \STATE norm\_entropy = entropy.mean() 
    % \PYCOMMENT{$[B, 1, 1]$} 
    
    \vspace{1em}
    
    \STATE \PYCOMMENT{obtain spatial enhancement parameters} \\
    A\_r, A\_t = $spatial\_by\_entropy$(norm\_entropy, $k_r$, $d_r$, $k_t$, $d_t$) 
    
    \PYCOMMENT{apply $\mathcal{A}_{r}$ and $\mathcal{A}_{t}$, get predictions} \\
    \STATE x\_spatial = $aug\_spatial$(x, A\_r, A\_t)
    \STATE p\_spatial = $f$(x\_spatial) 
    \STATE
    \PYCOMMENT{apply the same enhancements to p\_w}\\
    \STATE p\_w\_spatial = $aug\_spatial$(p\_w, A\_r, A\_t)	
    
    \vspace{1em}
    
    \PYCOMMENT{loss function} \\
    \STATE $criterion$ = nn.MSELoss()
    \STATE loss\_spatial = $criterion$(p\_spatial, p\_w\_spatial)
\ENDFOR

% \Statex

% \PYCOMMENT{Map entropy to rotation angle and translation degree} \\
% \Function{$\texttt{spatial\_by\_entropy}$}{entropy, $k_r$, $d_r$, $k_t$, $d_t$}
%     \State $A_r \gets \frac{k_r}{1 + \exp(d_r - H)} \cdot r_{\text{max}}$
%     \State $A_t \gets \frac{k_t}{1 + \exp(d_t - H)} \cdot t_{\text{max}}$
%     \State \Return $A_r, A_t$
% \EndFunction

\end{algorithmic}
\label{alg_pseudo}
% \vspace{-1.0em}
\end{algorithm}

% 引入
% This section begins with a concise definition of the SSSS problem in Subsection~\ref{preliminaries}. We then present a summary of the proposed ASAug in Subsection~\ref{adaaug}. Next, we detail our augmentation approach, known as dynamic mapping augmentation, in Subsection~\ref{DME}. Lastly, we introduce our pixel-level consistency learning in Subsection~\ref{CL}.

% 预先定义
\subsection{Preliminaries}
\label{preliminaries}
The SSSS task focuses on training a segmentation model that optimally adapts to data distributions with a scarcity of labeled data and a surplus of unlabeled data. The labeled dataset ${D}^l={\{({x}^l,{y}^l)\}}^M$ consists of $M$ examples with labels, while the unlabeled dataset ${D}^u={\{{x}^u\}}^N$ includes $N$ images without labels, where $N >> M$. Our objective is to optimize a model utilizing the mean-teacher (MT)\cite{tarvainen2017mean} strategy, in which the prediction from a teacher model is regarded as ground truth for weakly augmented unlabeled instances.
Initially, the student model undergoes training with labeled images, followed by optimization on unlabeled data. The teacher model begins as a copy of the student model and is subsequently refined using the EMA~\cite{tarvainen2017mean} protocol.
The training loss $\mathcal{L}$ can be minimized by utilizing both the labeled $\mathcal{L}^l$ and unlabeled $\mathcal{L}^u$ datasets. 
% The loss function $\mathcal{L}$ is comprised of two parts: $\mathcal{L}^{\mathrm{l}}$, representing the loss on the labeled data, and $\mathcal{L}^{\mathrm{u}}$, representing the loss on the unlabeled data.
\begin{align}
    \mathcal{L} = \mathcal{L}^l + \lambda \mathcal{L}^u,
\end{align}
where $\lambda$ represents the hyper-parameter that balances the supervised and unsupervised trade-offs. 
It can be set to a constant value \cite{sohn2020fixmatch,lai2021semi} or dynamically modified \cite{zhou2021c3,kong2023pruning} throughout the training phase. 

% 交叉熵损失详细
The supervised loss $\mathcal{L}^l$ can be expressed as:
\begin{align}
    \mathcal{L}^l=-\frac{1}{M}\sum_{l=1}^{M}\sum_{c=1}^Cy^{l}_{c}\log(p^{l}_{c}(\mathcal{T}_w(x^l);\theta))
\end{align}
where $p^{l}_{c}(\cdot;\theta)$ is the model's predicted probability that the sample $x^l$ belongs to category $c$. $C$ is the number of categories. $\mathcal{T}_w(\cdot)$ represents the weak augmentation, a commonly used method.
$y^{l}_{c}$ is the ground truth for sample $x^l$ with category label $c$ (one-hot).
% The unsupervised loss $\mathcal{L}^u$ can be computed differently according to the chosen method. Our implementation details will be addressed in subsequent sections.
% is the key to distinguishing between different semi-supervised methods, so it is presented in different forms in different methods.
% \begin{align}
%     \mathcal{L}^{\mathrm{u}}=-\frac1{{\mathrm{q}}}\sum_{u=1}^{{\mathrm{q}}}\sum_{c=1}^C\tilde{y}_{u,c}\log(p^{u}_{c}(x^u;\theta))
% \end{align}
% where $p^{l}_{c}(x^l;\theta)$ and $p^{u}_{c}(x^u;\theta)$ is the model's predicted probability that the sample $x^l$ and $x^u$ belongs to category $c$, respectively. 
% $C$ is the number of categories. 
% $y^{l}_{c}$ is the true corresponding to sample $x^l$ with category label (one-hot coding). 
% $\tilde{y}_{u,c}$ is the pseudo-label of $x^u$ (which can be either hard or soft) generated by the model, usually selected by the maximum probability category output by the model. 

% 介绍一下传统的弱到强的一致性正则化
As for the unsupervised loss $\mathcal{L}^u$, the key to SSL, is computed differently depending on the chosen method. 
Traditional weak-to-strong consistent regularization (WSCR) relies on strong and weak augmentation strategies to produce divergent predictions for the same input, obtaining the segmentation predictions $p^{u}_{s}=f(\mathcal{T}_s(x^u))$ for the student model on the strong augmentation $\mathcal{T}_s(\cdot)$ and $p^{u}_{w}=f(\mathcal{T}_w(x^u))$ for the teacher model on the weakly augmentation $\mathcal{T}_w(\cdot)$ to compute the loss, respectively:
\begin{align}
    \mathcal{L}^{\mathrm{consistency}}=-\frac1{{\mathrm{N}}}\sum_{u=1}^{{\mathrm{N}}}\sum_{c=1}^C p^{u}_{w,c}\log(p^{u}_{s,c})
\end{align}
where $\mathcal{T}_s(\cdot)$ denotes intensity enhancement transform applied to $x^u$ (e.g., brightness). This is also the main technique recently used for WSCR in SSSS and has shown effectiveness. 
In contrast, widely used spatial enhancements (e.g., rotations and translations) are usually ignored. 
Strong spatial augmentation can effectively improve the model robustness, therefore, we propose an adaptive spatial augmentation (ASAug), the details are described below.

% pascal classic 集结果
\begin{table*}[!t]
\vspace{-1.0em}
% \normalsize
\footnotesize
% \small
% \fontsize{10}{8}
\centering
\caption{Comparison with SOTA methods on the \textbf{Pascal} $classic$ val set (Indicator: $mIoU$). 
% The labeled images are selected from the classical VOC training set, which consists of 1464 images. 
Two crop sizes are tested.}
% \textasciitilde{} denotes the results we reproduced
\vspace{-0.5em}
\begin{tabular}{c|c|ccccc}
% \hline
\toprule[1pt]
\rowcolor[HTML]{ECF4FF} 
\textbf{Method} & 
\textbf{Size} & 
\textbf{1 / 16 ( 92 )} & 
\textbf{1 / 8 ( 183 )} & 
\textbf{1 / 4 ( 366 )} & 
\textbf{1 / 2 ( 732 )} & 
\textbf{Full ( 1464 )}  \\ 
% \hline
\midrule
Supervised           
&513$\times$513& 50.43 & 56.20 & 64.73 & 67.37 & 70.35 \\
% Mean Teacher\cite{tarvainen2017mean}   
% &321$\times$321& 52.72 & 58.93 & 65.92 & 69.54 & 72.42 \\
CutMix-Seg\cite{french2019semi}     
&513$\times$513& 55.58 & 63.20 & 68.36 & 69.84 & 76.54 \\
PseudoSeg\cite{zoupseudoseg}
&513$\times$513& 57.60 & 65.50 & 69.14 & 72.41 & —     \\
CPS\cite{chen2021semi}                  
&513$\times$513& 64.07 & 67.42 & 71.71 & 75.88 & —     \\
% PC2Seg\cite{zhong2021pixel}               
% &513$\times$513& 57.00 & 66.28 & 69.78 & 73.05 & 74.15 \\
PS-MT\cite{liu2022perturbed}                
&513$\times$513& 65.80 & 69.58 & 76.57 & 78.42 & 80.01 \\
GTA\cite{jin2022semi}                  
&513$\times$513& 70.00 & 73.20 & 75.60 & 78.40 & 80.50 \\
% RC2L\cite{zhang2022region}                 
% &513$\times$513& 65.30 & 68.90 & 72.20 & 77.10 & 79.30 \\
% PCR\cite{xu2022semi}                 
% &513$\times$513& 70.10 & 74.70 & 77.20 & 78.50 & 80.70 \\
AugSeg \cite{zhao2023augmentation}              
&513$\times$
513& 71.09 & 75.45 & 78.80 & 80.33 & 81.36 \\
CCVC \cite{wang2023conflict}                
&513$\times$
513& 70.20 & 74.40 & 77.40 & 79.10 & 80.50 \\ 
% \hline
\midrule
Allspark\cite{wang2024allspark}         
&513$\times$513
& 76.09 & 78.41 & 79.77 & 80.75 & 82.12 
\\
% Allspark\textasciitilde{}
% &513$\times$513& 73.74 & 75.76 & 80.71 & 80.39 & 82.03 \\
\rowcolor[HTML]{ECF4FF} 
\textbf{Allspark w/ ASAug}
&513$\times$513
& \textbf{77.29}  
&  \textbf{78.98} 
& \textbf{80.25}    
& \textbf{81.36}   
&  \textbf{83.25}  \\ 
% \bottomrule
\rowcolor[HTML]{ECF4FF} 
\textbf{
\begin{tabular}[c]{@{}c@{}} 
{\color[HTML]{3531FF} \textbf{$\uparrow $ $\bigtriangleup$ (\%) }}
\end{tabular}}
& \textbf{\color[HTML]{3531FF}  —}     
& \textbf{\color[HTML]{3531FF} $\uparrow $ 1.20}     
& \textbf{\color[HTML]{3531FF} $\uparrow $ 0.57}   
& \textbf{\color[HTML]{3531FF} $\uparrow $ 0.48} 
& \textbf{\color[HTML]{3531FF} $\uparrow $ 0.61} 
& \textbf{\color[HTML]{3531FF} $\uparrow $ 1.13} 
\\
% \hline
\midrule
CorrMatch\textasciitilde{}\cite{sun2024corrmatch}
&513$\times$513
& 76.41   & 77.83    & 79.57  & 80.66   & 81.93  \\
\rowcolor[HTML]{ECF4FF} 
\textbf{CorrMatch w/ ASAug} 
&513$\times$513
& 77.06    
& 78.75     
& 80.09      
& 81.03      
& 82.40      
\\ 
% \bottomrule
\rowcolor[HTML]{ECF4FF} 
\textbf{
\begin{tabular}[c]{@{}c@{}} 
{\color[HTML]{3531FF} \textbf{$\uparrow $ $\bigtriangleup$ (\%) }}
\end{tabular}}
& \textbf{\color[HTML]{3531FF}  —}     
& \textbf{\color[HTML]{3531FF} $\uparrow $ 0.65}     
& \textbf{\color[HTML]{3531FF} $\uparrow $ 0.92}   
& \textbf{\color[HTML]{3531FF} $\uparrow $ 0.52} 
& \textbf{\color[HTML]{3531FF} $\uparrow $ 0.37} 
& \textbf{\color[HTML]{3531FF} $\uparrow $ 0.47} 
\\
% \hline
\midrule[0.75pt]
\midrule[0.75pt]
Supervised           
&321$\times$321& 48.71 & 55.35 & 60.13 & 66.26 & 70.01 \\
Mean Teacher\cite{tarvainen2017mean}   
&321$\times$321& 52.72 & 58.93 & 65.92 & 69.54 & 72.42 \\
UniMatch \cite{yang2023revisiting}           
&321$\times$321& 75.20 & 77.20 & 78.80 & 79.90 & 81.20 \\
% \hline
\midrule
Allspark\textasciitilde{}\cite{wang2024allspark}   
&321$\times$321
& 72.36 
& 76.31 
& 77.07 
& 79.56 
& 80.62 \\
\rowcolor[HTML]{ECF4FF} 
\textbf{Allspark w/ ASAug}
&321$\times$321& 
76.52 
& 78.76 
& 79.92 
& \textbf{81.11} 
& \textbf{83.00} 
\\
% \bottomrule
\rowcolor[HTML]{ECF4FF} 
\textbf{
\begin{tabular}[c]{@{}c@{}} 
{\color[HTML]{3531FF} \textbf{$\uparrow $ $\bigtriangleup$ (\%) }}
\end{tabular}}
& \textbf{\color[HTML]{3531FF}  —}     
& \textbf{\color[HTML]{3531FF} $\uparrow $ 4.16}     
& \textbf{\color[HTML]{3531FF} $\uparrow $ 2.45}   
& \textbf{\color[HTML]{3531FF} $\uparrow $ 2.85} 
& \textbf{\color[HTML]{3531FF} $\uparrow $ 1.55} 
& \textbf{\color[HTML]{3531FF} $\uparrow $ 2.38} 
\\
% \hline
\midrule
CorrMatch\cite{sun2024corrmatch}
&321$\times$321& 76.40 & 78.50 & 79.40 & 80.60 & 81.80 \\
% CorrMatch\textasciitilde{}\cite{sun2024corrmatch}
% &321$\times$321& 75.31 & 76.98 & 79.35 & 80.01 & 81.18 \\
\rowcolor[HTML]{ECF4FF} 
\textbf{CorrMatch w/ ASAug }
&321$\times$321
& \textbf{76.91} 
& \textbf{79.31}
& \textbf{79.98} 
& 80.94 
& 82.28 
\\
% \bottomrule
\rowcolor[HTML]{ECF4FF} 
\textbf{
\begin{tabular}[c]{@{}c@{}} 
{\color[HTML]{3531FF} \textbf{$\uparrow $ $\bigtriangleup$ (\%) }}
\end{tabular}}
& \textbf{\color[HTML]{3531FF}  —}     
& \textbf{\color[HTML]{3531FF} $\uparrow $ 0.51}     
& \textbf{\color[HTML]{3531FF} $\uparrow $ 0.81}   
& \textbf{\color[HTML]{3531FF} $\uparrow $ 0.58} 
& \textbf{\color[HTML]{3531FF} $\uparrow $ 0.34} 
& \textbf{\color[HTML]{3531FF} $\uparrow $ 0.48} 
\\
% \hline
\bottomrule[1pt]
\end{tabular}
\begin{tablenotes}
\small     
% \normalsize
\item[1] \textasciitilde{} denotes the results we reproduced. 
% \item [2] 
We split the table with the different crop sizes.
\end{tablenotes}   
\label{tab: pascal_classic}
\vspace{-1.0em}
\end{table*}

\subsection{Adaptive Spatial Augmentation}
\label{adaaug}
% Recent weak-to-strong consistency regularization methods for SSSS mainly use intensity-based augmentations and show effectiveness. In contrast, widely used strong spatial augmentation, such as rotation and translation, is usually ignored. We give the details of our adaptive spatial augmentation as follows.
% 
In particular, we substitute intensity-focused robust augmentations with strong spatial augmentation $\mathcal{T}_{ss}$, which involves both rotational $\mathcal{A}_r$ and translational $\mathcal{A}_t$ transformations. 
Compared to pixel level modifications like jittering and brightness change, $\mathcal{A}_r$ and $\mathcal{A}_t$ import more vibrant disturbance to unlabeled training samples.
\begin{itemize}
    \item \textbf{Rotation $\mathcal{A}_r$}: Real-world objects and scenes are viewed from multiple perspectives. By rotating images, we simulate these different angles, helping the model identify patterns and characteristics across orientations. In ASAug, we adjust the rotation within a restricted angle.
    % Rotation angle is dynamically adjusted. 
    % When $\mathcal{H}$ is higher, the rotation angle increases to cope with the uncertainty in different directions, thus enhancing the robustness of the model to spatial transformations. 
    % It has been shown that a reasonable rotation transformation helps to enhance the orientation invariance of the model \cite{}. 
    \item \textbf{Translation $\mathcal{A}_t$}: In reality, cameras frequently adjust their positions horizontally or vertically to capture scenes from different angles. Translating images allows us to mimic these camera movements, which helps the model identify patterns and features in varied locations. In ASAug, we achieve this by shifting an image sideways or up and down by a specified pixel count.
    % Also adjusts the translation amplitude based on $\mathcal{H}$. Higher entropy samples correspond to larger translation amplitudes to enhance the model's ability to learn consistently across different spatial locations. 
    % By training on diverse locations, the model can gradually learn to maintain consistent predictions in the presence of changing object positions \cite{}. 
\end{itemize}

To tackle the challenges posed by increased uncertainty biases, we propose the adaptive spatial augmentation that modifies the augmentation intensity based on the sample's quality. 
In WSCR works, the prediction of weak augmented image $\mathcal{T}_w(x^u)$  from the teacher model is treated as ground truth for $x^u$. 
We use the entropy of this prediction $p^u(\mathcal{T}_w(x^u))$ as the measurement for instance reliability. 
Specifically, we assess the prediction's entropy $\mathcal{H}$ as an indicator of reliability and adjust the level of strong spatial augmentation $\mathcal{T}_{ss}(\mathcal{A}_r,\mathcal{A}_t)$ accordingly.
For the predicted output probability $p^u(\cdot)$, the $\mathcal{H}$ is defined as follows:
\begin{equation}
\mathcal{H}=-\sum_{c=1}^Cp^{u}_{c}(\mathcal{T}_w(x^u))\log(p^{u}_{c}(\mathcal{T}_w(x^u))
\end{equation}
where $C$ denotes the number of categories, $p^{u}_c(\mathcal{T}_w(x^u))$ is the predicted  probability of sample $x^u$ on category $c$. 
The higher the value of $\mathcal{H}$, the more uncertain the model is about its prediction for that particular sample, and conversely.

To fine-tune the level of augmentation distortion, we introduce an entropy-based adaptive weight, utilizing information entropy as a measure of sample reliability. This means that the rotation angle and the degree of translation are modulated according to the entropy value of the weakly augmented prediction, denoted by $\mathcal{H}$. 
EAW dynamically adjusts the strength of spatial transformations. 
Samples with high entropy, where the model exhibits greater uncertainty, require more significant spatial transformations to explore a broader feature space. Conversely, samples with low entropy, where the model demonstrates greater certainty, are better suited for smaller transformations, thus preserving stable features. 
This EAW approach enables a smoother adjustment of augmentation across samples, enhancing the model's capacity and resilience to spatial transformations.

% In the initial design, we used the standard Sigmoid function to smooth the mapping of entropy values.
% \begin{equation}
% \mathrm{Sigmoid}(\mathcal{H})=\frac1{1+e^{-\mathcal{H}}}
% \end{equation}
% However, when applied specifically to rotation and translation amplitudes, the output is less flexible in controlling low-entropy data. Specifically, since the entropy range is limited to $[0, 3.0]$, the output is close to $0.5$ at $\mathcal{H}=0$, which leads to high intensity of geometric transformation for low entropy data (entropy value close to zero), and fails to meet the stability requirements of low entropy samples effectively.
% We improved the mapping function by introducing an offset $d$ to adjust the entropy value of the input, which better meets our need for differential enhancement of different samples. 
% The improved form of the mapping function $\sigma$ is as follows:
% \begin{equation}
% \sigma(\mathcal{H})=\frac1{1+e^{-(\mathcal{H}-d)}}
% \end{equation}
We define our EAW used in adaptive spatial augmentation $\mathcal{A}_r(\mathcal{H})$ and $\mathcal{A}_t(\mathcal{H})$ as follows:
\begin{equation}
% \mathcal{A}_r(\mathcal{H})=r_{\max}\cdot\sigma(k_r\cdot\frac1{1+e^{-(\mathcal{H}-d_r)}})
\mathcal{A}_r(\mathcal{H})=r_{\max} \cdot {k_r}\cdot({1+e^{d_r-\mathcal{H}})^{-1}}
\end{equation}
% \vspace{-0.9em}
\begin{equation}
\mathcal{A}_t(\mathcal{H})=t_{\max} \cdot {k_t}\cdot({1+e^{d_t-\mathcal{H}})^{-1}}
\end{equation}
where $k_r, k_t$ are the scaling parameters, $r_{max}$ and $t_{max}$ denote the maximum rotation angle ($^{\circ}$) and maximum translation ratio ($\times 100\%$), respectively.
$d_r$ and $d_t$ are the offsets associated with the two spatial transformations to fine-tune the input entropy values.
We combine $\mathcal{A}_r$ and $\mathcal{A}_t$ to form the strong spatial augmentation $\mathcal{T}_{ss}(\mathcal{A}_r,\mathcal{A}_t)$.

When $\mathcal{H}$ is relatively small, the mapping outcome is likewise reduced, which decreases the magnitude of rotation and translation, thereby maintaining the stability of the applied enhancement.

% 损失函数
\subsection{Pixel-level Consistency Learning}\label{CL}
When applying spatial transformations, the structure of the image changes, and so does the position of the mask. 
To ensure that the loss is a correct measure of the difference in prediction before and after enhancement, we apply the same spatial transformations to the weakly enhanced forecasts as we do to the strongly enhanced ones to align their spatial structure.
We adopt the mean squared error (MSE) as the metric of consistency loss.
Compared to other losses, like cross-entropy loss, MSE is smoother and more robust in measuring the variance of consecutive predicted values. 
It is particularly suitable for calculating prediction consistency that does not depend on category labels under geometric enhancement. 

The pixel-level consistency loss $\mathcal{L}^u$  for unlabelled samples is defined as follows:
\begin{equation}
\begin{aligned}
\mathcal{L}^{u} 
=\frac{1}{N} \sum_{u=1}^{N}\frac{1}{HW}  \sum_{i=1}^H\sum_{j=1}^W  
\left(\mathcal{T}_{ss}(p(\mathcal{T}_{w}(x^u_{i,j})))- 
p(\mathcal{T}_{ss}(x^u_{i,j}))\right)^2
\end{aligned}
\end{equation}
where $H$ and $W$ are the height and width of the image, respectively. 
The difference in predicted values at each pixel location $(i,j)$ is squared summed, and averaged to obtain a pixel-level MSE for that image. MSE directly measures spatially aligned pixel differences without sensitivity to the mask structure, improving the model's robustness to the augmentation transform.

% Pascal blender 数据集结果
\begin{table}[!t]
\vspace{-1.0em}
% \small
\footnotesize
% \fontsize{6}{10}
\centering
\caption{Comparison with SOTA methods on the \textbf{Pascal} $blender$ val set (Indicator: $mIoU$). The split table reports results by following U2PL~\cite{wang2022semi} configurations.}
% The labeled images are selected from the augmented VOC training set, which consists of 10,852 images 
% .}
% \vspace{-1.0em}
\begin{tabular}{c|c|cccc}
% \hline
\toprule[1pt]
\rowcolor[HTML]{ECF4FF} 
\textbf{Method} &
  \textbf{Size} &
  \textbf{\begin{tabular}[c]{@{}c@{}}1/16\\(662)\end{tabular}} &
  \textbf{\begin{tabular}[c]{@{}c@{}}1/8\\(1323)\end{tabular}} &
  \textbf{\begin{tabular}[c]{@{}c@{}}1/4\\(2646)\end{tabular}} &
  \textbf{\begin{tabular}[c]{@{}c@{}}1/2\\(5291)\end{tabular}} \\ 
  \midrule
  % \hline
Supervised  &321$\times$321
& 65.58          & 70.37          & 72.46          & 73.61          \\
CAC\cite{lai2021semi}  &321$\times$321
& 72.40          & 74.60          & 74.30          & —               \\
UniMatch\cite{yang2023revisiting}    &321$\times$321
& 76.50          & 77.00          & 77.20          & —               \\
CutMix-Seg\cite{french2019semi}      &513$\times$513
& 72.56          & 72.69          & 74.25          & 75.89          \\
CPS\cite{chen2021semi}   &513$\times$513
& 72.18          & 75.83          & 77.55          & 78.64          \\
PS-MT\cite{liu2022perturbed}   &513$\times$513
& 75.50          & 78.20          & 78.72          & 79.76          \\
ESL\cite{ma2023enhanced}     &513$\times$513
& 76.36          & 78.57          & 79.02          & 79.98          \\ 
\midrule
% \hline
Allspark\textasciitilde{}\cite{wang2024allspark}&321$\times$321
& 76.89          & 78.18          & 80.02          & 79.86          \\
\rowcolor[HTML]{ECF4FF} 
\textbf{\begin{tabular}[c]{@{}c@{}}
% Allspark\\ 
% w/ AdaAug
w/ ASAug
\end{tabular}}   &321$\times$321
& 77.60 
& \textbf{80.22} 
& \textbf{80.97} 
& \textbf{80.37} 
\\ 
% \bottomrule
\rowcolor[HTML]{ECF4FF} 
\textbf{
\begin{tabular}[c]{@{}c@{}} 
{\color[HTML]{3531FF} \textbf{$\uparrow $ $\bigtriangleup$ (\%) }}
\end{tabular}}
& —
& \textbf{\color[HTML]{3531FF} $\uparrow $ 0.71}     
& \textbf{\color[HTML]{3531FF} $\uparrow $ 2.04}   
& \textbf{\color[HTML]{3531FF} $\uparrow $ 0.95} 
& \textbf{\color[HTML]{3531FF} $\uparrow $ 0.51} 
\\
\midrule
% \hline
CorrMatch\cite{sun2024corrmatch}        
&321$\times$321
& 77.60  
& 77.80    
& 78.30          
& —               \\
\rowcolor[HTML]{ECF4FF} 
\textbf{\begin{tabular}[c]{@{}c@{}}
% CorrMatch\\ 
% w/ AdaAug
w/ ASAug
\end{tabular}}  &321$\times$321
& \textbf{78.52} 
& 78.27 
& 78.87 
& —     
\\ 
% \bottomrule
\rowcolor[HTML]{ECF4FF} 
\textbf{
\begin{tabular}[c]{@{}c@{}} 
{\color[HTML]{3531FF} \textbf{$\uparrow $ $\bigtriangleup$ (\%) }}
\end{tabular}}
& —
& \textbf{\color[HTML]{3531FF} $\uparrow $ 0.92}     
& \textbf{\color[HTML]{3531FF} $\uparrow $ 0.47}   
& \textbf{\color[HTML]{3531FF} $\uparrow $ 0.57} 
& \textbf{\color[HTML]{3531FF} —} 
\\
\midrule[0.75pt]
\midrule[0.75pt]
% \hline
U2PL$^{\dagger}$\cite{wang2022semi}    &513$\times$513
& 74.40          & 77.60          & 78.70          & —               \\
AugSeg$^{\dagger}$\cite{zhao2023augmentation}   &513$\times$513
& 77.01          & 78.20          & 78.82          & —               \\ 
CCVC$^{\dagger}$\cite{wang2023conflict}   &513$\times$513
& 77.20          & 78.40          & 79.00          & —               \\
iMAS$^{\dagger}$\cite{zhao2023instance} &513$\times$513
& 77.20          & 78.40          & 79.30          & —               \\

\midrule
% \hline
Allspark$^{\dagger}$\textasciitilde{}  &321$\times$321
& 80.44  
& 81.00               
&79.34 
&78.91                \\
\rowcolor[HTML]{ECF4FF} 
\textbf{\begin{tabular}[c]{@{}c@{}}
% Allspark\\ 
w/ ASAug$^{\dagger}$
% w/ AdaAug$^{\dagger}$
% 000000 黑色
% FE0000 红色
\end{tabular}}  &321$\times$321
& \textbf{\color[HTML]{000000} 81.25} 
& \textbf{82.39} 
& \textbf{82.70} 
& \textbf{80.88} 
\\ 
  % \bottomrule
  \rowcolor[HTML]{ECF4FF} 
  \textbf{
\begin{tabular}[c]{@{}c@{}} 
{\color[HTML]{3531FF} \textbf{$\uparrow $ $\bigtriangleup$ (\%) }}
\end{tabular}}
& —
& \textbf{\color[HTML]{3531FF} $\uparrow $ 0.81}     
& \textbf{\color[HTML]{3531FF} $\uparrow $ 1.39}   
& \textbf{\color[HTML]{3531FF} $\uparrow $ 3.36} 
& \textbf{\color[HTML]{3531FF} $\uparrow $ 1.97} 
\\
\midrule
% \hline
CorrMatch$^{\dagger}$\textasciitilde{}  
&321$\times$321
& 78.44  & 80.64   & 79.08   & 78.14          \\
\rowcolor[HTML]{ECF4FF} 
\textbf{\begin{tabular}[c]{@{}c@{}}
% CorrMatch\\ 
% w/ AdaAug$^{\dagger}$
w/ ASAug$^{\dagger}$
\end{tabular}} &321$\times$321
& 79.55 
& 81.77 
& 79.85 
& 78.69 
\\ 
  % \bottomrule
  \rowcolor[HTML]{ECF4FF} 
  \textbf{
\begin{tabular}[c]{@{}c@{}} 
{\color[HTML]{3531FF} \textbf{$\uparrow $ $\bigtriangleup$ (\%) }}
\end{tabular}}
& —
& \textbf{\color[HTML]{3531FF} $\uparrow $ 1.11}     
& \textbf{\color[HTML]{3531FF} $\uparrow $ 1.13}   
& \textbf{\color[HTML]{3531FF} $\uparrow $ 0.77} 
& \textbf{\color[HTML]{3531FF} $\uparrow $ 0.55} 
\\
% \hline
\bottomrule[1pt]
\end{tabular}
\begin{tablenotes}
% \small     
% \normalsize
\footnotesize
\item[1] \textasciitilde{} denotes the results we reproduced. 
\item[2] $^{\dagger}$ means the same split as U2PL\cite{wang2022semi}, which are contained in split table.
% \item [3] Our experiments were all conducted under size=321$\times$321, and all methods in the table except $^*$ have larger training volumes than ours (size=513$\times$513).
\end{tablenotes}   
\label{tab:pascal_aug}
% \vspace{-1.0em}
\end{table}

% 消融实验结果
\begin{table}[!t]
% \vspace{-0.5em}
% \footnotesize
\small
% \normalsize
\centering
\caption{Ablation study on our ASAug. $\mathcal{A}_r$ and $\mathcal{A}_t$ represent the two main augmentation operations, respectively. Improvements to the baseline are highlighted in {\color[HTML]{3531FF} blue}. 
% (Based on CorrMatch~\cite{sun2024corrmatch}).
}
% \vspace{-0.5em}
% \vspace{-1.0em}
\begin{tabular}{ccccc|cc}
% \hline
\toprule[1pt]
\rowcolor[HTML]{ECF4FF} 
\multicolumn{5}{c|}{\cellcolor[HTML]{ECF4FF}\textbf{ASAug}} 
& \multicolumn{2}{c}{\cellcolor[HTML]{ECF4FF}\textbf{mIoU(\%)}} \\ 
% \hline
\midrule
% MT
&\textbf{$\mathcal{A}_r$} 
& 
&\textbf{$\mathcal{A}_t$} 
& 
& 
\textbf{732} & \textbf{366} 
\\
\midrule
% \hline
% &    
% & 
% & 66.26 {\color[HTML]{3531FF} (supervised)} 
% & 60.13 {\color[HTML]{3531FF} (supervised)}
% \\
% \checkmark  
&    
& 
& 
& 
& 77.85  
& 76.77  
\\
% \checkmark  
& \checkmark  
&    
& 
& 
& 78.71 {\color[HTML]{3531FF} ( $\uparrow $ 0.86 )}     
& 78.04 {\color[HTML]{3531FF} ( $\uparrow $ 1.27 )}      
\\
% \checkmark  
&    
& 
& \checkmark  
& 
& 78.18 {\color[HTML]{3531FF} ( $\uparrow $ 0.33 )}  
& 77.30 {\color[HTML]{3531FF} ( $\uparrow $ 0.53 )}      
\\ 
\midrule
% \hline
\rowcolor[HTML]{ECF4FF} 
% \checkmark  
& \checkmark  
& 
& \checkmark  
& 
% FE0000
& \textbf{80.94 }
\textbf{\color[HTML]{3531FF} ( $\uparrow $ 3.09 )} 
& \textbf{79.98 }
\textbf{\color[HTML]{3531FF} ( $\uparrow $ 3.21 )} \\ 
\bottomrule[1pt]
% \hline
\end{tabular}
\label{ablation_component}
\vspace{-0.5em}
\end{table}

\section{Experiments}
% 引入文字
This section starts with a thorough explanation of our experimental setup in subsection~\ref{details}. Following that, subsection~\ref{comparison} is dedicated to assessing ASAug in comparison with leading algorithms on two popular SSSS benchmarks. Next, in subsection~\ref{ablation}, we perform ablation studies to further validate and confirm the robustness of ASAug. Finally, subsection~\ref{QA} offers a qualitative analysis, supported by visual results of Cityscapes.

% 设置
\subsection{Implementation Details}\label{details}
% \vspace{-0.5em}
\textbf{Datasets.}
We investigate the effects of ASAug on three benchmark segmentation datasets: Pascal VOC 2012 \cite{everingham2015pascal}, Cityscapes \cite{cordts2016cityscapes}, and COCO \cite{lin2014microsoft}. 
The Pascal VOC 2012 dataset features $21$ semantic categories and is divided into $classic$ and $blender$ subsets. The $classic$ subset consists of $1,464$ images with extensive labels for training and $1,449$ images for validation. Conversely, the $blender$ subset, as described in \cite{wang2022semi,yang2022st++}, includes lower-resolution, roughly annotated images from the Segmentation Boundary Dataset (SBD) \cite{hariharan2011semantic}, thus enlarging the training pool to $10,582$ images. We also apply the same settings as U2PL \cite{wang2022semi} to evaluate ASAug on the $blender$ subset. 
The Cityscapes dataset, which encompasses $19$ semantic categories within urban contexts, offers $2,975$ training images with precise annotations and $500$ images for validation.
The COCO dataset contains $80$ categories and provides $118,000$ training images and $5,000$ validation images covering indoor and outdoor scenes.

% Cityscapse 数据集结果
\begin{table}[!t]
\vspace{-1.0em}
\footnotesize
% \small
% \normalsize
% \fontsize{8}{10}
\centering
\caption{Comparison with SOTAs on the \textbf{Cityscapes} val set (Indicator: $mIoU$). Images are cropped to 801$\times$801.}
% \vspace{-1.0em}
\begin{tabular}{c|cccc}
% \hline
\toprule[1pt]
\rowcolor[HTML]{ECF4FF} 
\textbf{Method} &
  \textbf{\begin{tabular}[c]{@{}c@{}}1/16(186) \end{tabular}} &
  \textbf{\begin{tabular}[c]{@{}c@{}}1/8(372) \end{tabular}} &
  \textbf{\begin{tabular}[c]{@{}c@{}}1/4(744)  \end{tabular}} &
  \textbf{\begin{tabular}[c]{@{}c@{}}1/2(1488)  \end{tabular}} 
  \\ 
\midrule
% \hline
Supervised    
& 64.39 & 72.14 & 74.83 & 77.93 \\
CPS\cite{chen2021semi}                                   
& 69.78 & 74.31 & 74.58 & 76.81 \\
AEL \cite{hu2021semi}
& 74.45 & 75.55 & 77.48 & 79.01 \\
PS-MT\cite{liu2022perturbed}
&  —    & 76.90 & 77.60 & 79.10 \\
U2PL\cite{wang2022semi}
& 70.30 & 73.37 & 76.47 & 79.05 \\
CCVC\cite{wang2023conflict}
& 74.90 & 76.40 & 77.30 &  —     \\
AugSeg\cite{zhao2023augmentation}     
& 75.22 & 77.82 & 79.56 & 80.43 \\
DGCL\cite{wang2023hunting}
& 73.18 & 77.29 & 78.48 & 80.71 \\
UniMatch\cite{yang2023revisiting} 
& 76.60 & 77.90 & 79.20 & 79.50 \\
ESL\cite{ma2023enhanced}
& 75.12 & 77.15 & 78.93 & 80.46 \\
% LogicDiag                                                         & 76.83 & 78.00 & 80.21 & 81.25 \\
CFCG\cite{li2023cfcg}
& 77.28 & 79.09 & 80.07 & 80.59 \\ 
\midrule[0.75pt]
% \hline
Allspark\cite{wang2024allspark}
& 78.33 & 79.24 & 80.56 & 81.39 \\
% Allspark\textasciitilde{}\cite{wang2024allspark}
% &       &       &       &       \\
\rowcolor[HTML]{ECF4FF} 
\textbf{\begin{tabular}[c]{@{}c@{}}
% Allspark \\
w/ ASAug
\end{tabular}}      
% &\textbf{{\fontsize{10}{10}\selectfont {}}}       
% &\textbf{{\fontsize{10}{10}\selectfont {}}}       
% &\textbf{{\fontsize{10}{10}\selectfont {81.06}}}       
% &\textbf{{\fontsize{10}{10}\selectfont {81.95}}} 
&\textbf{78.91}
&\textbf{79.68}
&\textbf{81.06}
&\textbf{81.95}
\\ 
% \bottomrule
\rowcolor[HTML]{ECF4FF} 
\textbf{
\begin{tabular}[c]{@{}c@{}} 
{\color[HTML]{3531FF} \textbf{$\uparrow $ $\bigtriangleup$ (\%) }}
\end{tabular}}
& \textbf{\color[HTML]{3531FF} $\uparrow $ 0.58}     
& \textbf{\color[HTML]{3531FF} $\uparrow $ 0.44}   
& \textbf{\color[HTML]{3531FF} $\uparrow $ 0.50} 
& \textbf{\color[HTML]{3531FF} $\uparrow $ 0.56} 
\\
% \hline
\midrule
CorrMatch\cite{sun2024corrmatch}
& 77.30 & 78.50 & 79.40 & 80.40 \\
% CorrMatch\textasciitilde{}\cite{sun2024corrmatch} 
% &       &       &       &       \\
\rowcolor[HTML]{ECF4FF} 
\textbf{\begin{tabular}[c]{@{}c@{}}
% CorrMatch \\
w/ ASAug
\end{tabular}} 
% &\textbf{{\fontsize{10}{10}\selectfont {}}}      
% &\textbf{{\fontsize{10}{10}\selectfont {}}}      
% &\textbf{{\fontsize{10}{10}\selectfont {79.86}}}
% &\textbf{{\fontsize{10}{10}\selectfont {80.98}}} 
&77.97
&79.22
&79.86
&80.98
\\ 
% \bottomrule
\rowcolor[HTML]{ECF4FF} 
\textbf{
\begin{tabular}[c]{@{}c@{}} 
{\color[HTML]{3531FF} \textbf{$\uparrow $ $\bigtriangleup$ (\%) }}
\end{tabular}}
& \textbf{\color[HTML]{3531FF} $\uparrow $ 0.67}     
& \textbf{\color[HTML]{3531FF} $\uparrow $ 0.72}   
& \textbf{\color[HTML]{3531FF} $\uparrow $ 0.46} 
& \textbf{\color[HTML]{3531FF} $\uparrow $ 0.58} 
\\
% &       &       &       &       \\ 
% \hline
\bottomrule[1pt]
\end{tabular}
\label{tab: city}
% \vspace{-1.0em}
\end{table}

% CoCo 数据集结果
\begin{table}[!t]
% \vspace{-0.5em}
\centering
\footnotesize
% \scriptsize
\caption{Comparison with SOTAs on the \textbf{COCO} val set (Indicator: $mIoU$). Images are cropped as 513$\times$513. Corrmatch\cite{sun2024corrmatch} did not report the COCO results.}
% \vspace{-1.0em}
\begin{tabular}{c|ccccc}
% \hline
\toprule[1pt]
\rowcolor[HTML]{ECF4FF} 
\textbf{Method} &
  \textbf{\begin{tabular}[c]{@{}c@{}}1/512\\ (232)\end{tabular}} &
  \textbf{\begin{tabular}[c]{@{}c@{}}1/256\\ (463)\end{tabular}} &
  \textbf{\begin{tabular}[c]{@{}c@{}}1/128\\ (925)\end{tabular}} &
  \textbf{\begin{tabular}[c]{@{}c@{}}1/64\\ (1849)\end{tabular}} &
  \textbf{\begin{tabular}[c]{@{}c@{}}1/32\\ (3697)\end{tabular}} \\ 
  % \hline
  \midrule
Supervised  
& 22.94 & 27.96 & 33.60 & 37.80 & 42.24  \\
PseudoSeg\cite{zoupseudoseg}
& 29.78 & 37.11 & 39.11 & 41.75 & 43.64  \\
PC2Seg\cite{zhong2021pixel}
& 29.94 & 37.53 & 40.12 & 43.67 & 46.05  \\
MKD\cite{yuan2023semi}
& 30.24 & 38.04 & 42.32 & 45.50 & 47.25  \\
UniMatch\cite{yang2023revisiting}
& 31.90 & 38.90 & 44.40 & 48.20 & 49.80  \\
LogicDiag\cite{liang2023logic}
& 33.07 & 40.28 & 45.35 & 48.83 & 50.51  \\
S4Former\cite{hu2024training}
& 35.20 
& 43.10 
& 46.90 & — & —      \\
BRPG\cite{dong2024boundary}
& —     & —     & 41.73 & 45.91 & 50.55  \\ 
% \hline
\midrule
Allspark\cite{wang2024allspark} 
& 34.10 & 41.65 & 45.48 & 49.56 & 51.72~ \\
\rowcolor[HTML]{ECF4FF} 
\textbf{\begin{tabular}[c]{@{}c@{}} 
w/ASAug
\end{tabular}}
&  \textbf{35.78} &
  \textbf{43.28} &
  \textbf{48.42} &
  \textbf{52.31} &
  \textbf{55.25} \\
  % \bottomrule
  \rowcolor[HTML]{ECF4FF} 
  \textbf{
\begin{tabular}[c]{@{}c@{}} 
{\color[HTML]{3531FF} \textbf{$\uparrow $ $\bigtriangleup$ (\%) }}
\end{tabular}}
& \textbf{\color[HTML]{3531FF} $\uparrow $ 1.68} 
& \textbf{\color[HTML]{3531FF} $\uparrow $ 1.63}     
& \textbf{\color[HTML]{3531FF} $\uparrow $ 2.94}   
& \textbf{\color[HTML]{3531FF} $\uparrow $ 2.75} 
& \textbf{\color[HTML]{3531FF} $\uparrow $ 3.53} 
\\
\bottomrule[1pt]
% \hline
\end{tabular}
\begin{tablenotes}
% \small     
% \normalsize
\footnotesize
% \scriptsize
\item[1] \textasciitilde{} denotes the results we reproduced, which are missing in the paper. 
\end{tablenotes}   
\vspace{-1.0em}
\label{tab: coco}
\end{table}

% 实验细节
% \
% \newline
\textbf{Parameter Settings.}
Following the conventions of most prior SSSS approaches~\cite{yang2022st++,zhao2023augmentation,yang2023revisiting}, we evaluated our ASAug technique on both traditional CNNs and ViT structures. 
Specifically, for the CNN-based CorrMatch\cite{sun2024corrmatch}, we utilized DeepLabV3+~\cite{chen2017rethinking} integrated with ResNet-101~\cite{he2016deep}, pre-trained on ImageNet~\cite{deng2009imagenet}, and for the ViT-based Allspark\cite{wang2024allspark}, we employed SegFormer-B5~\cite{xie2021segformer}. 
To evaluate ASAug's performance, SGD was used as the optimizer with a polynomial decay learning rate policy: $lr_{init} \cdot \left ( 1-\frac{iter}{total\_iters} \right ) ^{power}$, where $lr_{init}$ is the starting learning rate, $iter$ is the current iteration number, $total\_iters$ is the total number of iterations, with the power and weight decay set at $0.9$ and $1e-4$, accordingly. Loss weight $\lambda$ is typically set to 0.5. 
For PASCAL, we began with a learning rate of $0.001$, utilized crop sizes of $321\times321$ or $513\times513$, had a batch size of $16$, and trained for 80 epochs. 
For Cityscapes, the starting learning rate was $0.005$, employed a crop size of $801\times801$, used the same batch size, and trained for 240 epochs. All tests utilized the PyTorch deep learning framework and were executed on $4\times$ NVIDIA V100 and $2\times$ A800 GPUs for distinct numerical analyses. 
For COCO, we use the same initial learning rate and batch size as PASCAL, utilized crop sizes of $513\times513$, and train for 10 epochs. 
Regarding data augmentation, the parameters were set to $r_{max}=180$, $t_{max}=0.5$. For PASCAL, parameters were $d_r=1$, $d_t=1$, $k_r=11$, $k_t=7$. Meanwhile, for Cityscapes, some parameters were adjusted to $d_r=0.5$, $d_t=0.5$, $k_r=5.5$, and $k_t=3$ based on the specific context.

\textbf{Evaluations.}
In this experiment, we employ the mean intersection over union ($mIoU$) as our evaluation metric, which is a typical evaluation standard in SSSS. This metric is effective even when dealing with imbalanced classes, a frequent issue in pixel-level annotation tasks. Following the methodologies of CPS~\cite{chen2021semi} and U2PL~\cite{wang2022semi}, we also utilize a sliding evaluation approach to assess model performance on the Cityscapes~\cite{cordts2016cityscapes} dataset.

% 折现消融图
\begin{figure}[!t]
    \vspace{-0.5em}
    \centering
    \includegraphics[width=1.0\linewidth]{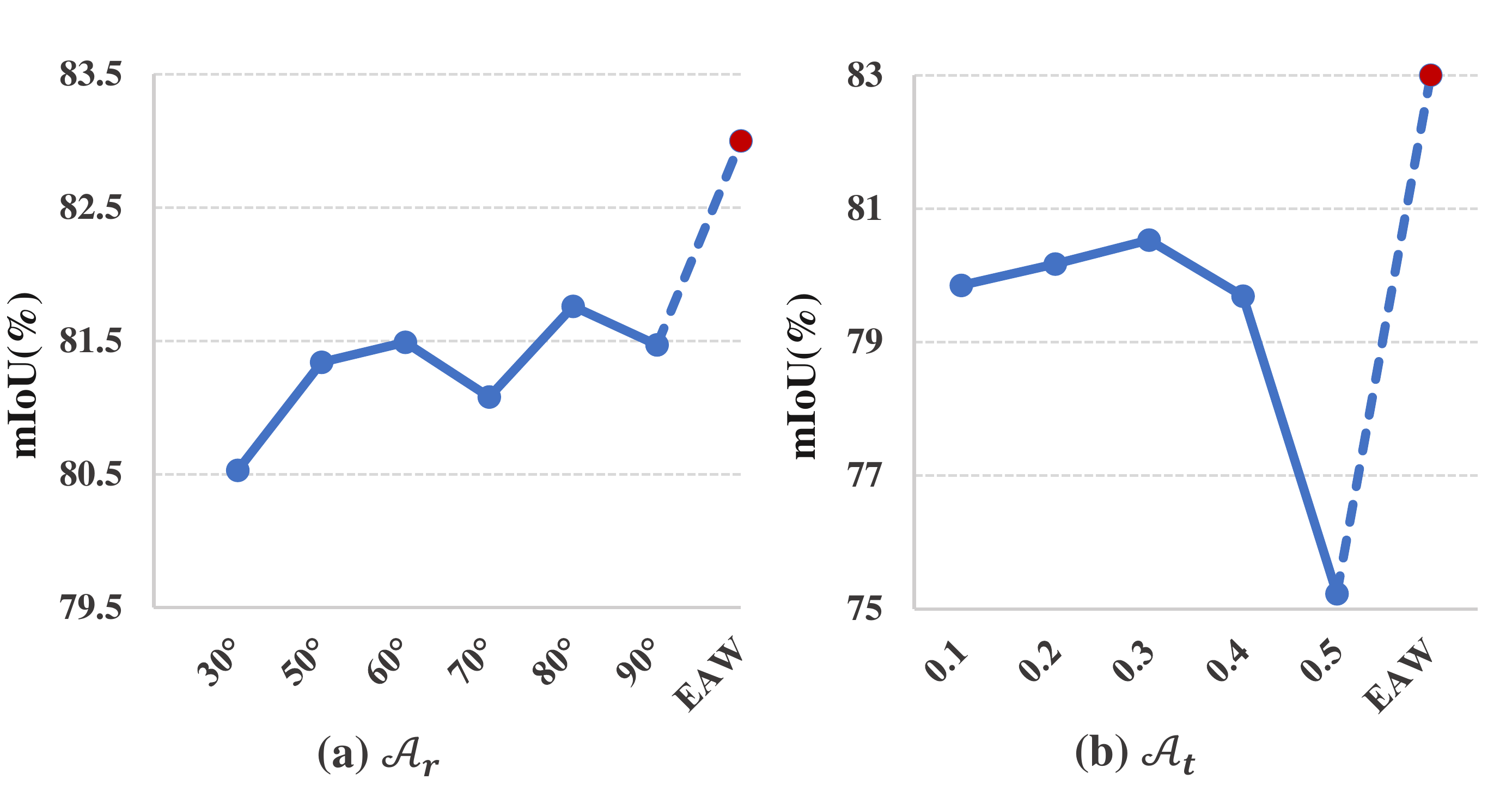}
    \vspace{-1.5em}
    \caption{Compare EAW with direct spatial augmentations. (a) EAW vs. fixed rotation angle, (b) EAW vs. same translation ratio (Based on Allspark\cite{wang2024allspark}).}
    \label{fig_ablation_m}
    \vspace{-1.0em}
\end{figure}

% 方法对比
\subsection{Comparison with SOTA Methods}
% \vspace{-0.5em}
\label{comparison}
Since our ASAug can be used as a pluggable module, we show its effectiveness by replacing the original strong augmentation with our method. In particular, we present the results of ASAug when combined with the highly competitive Allspark~\cite{wang2024allspark} and CorrMatch~\cite{sun2024corrmatch}, comparing its performance to leading methods in both datasets using the diverse partition strategies.

\textbf{Pascal VOC 2012 classic set.}
Table~\ref{tab: pascal_classic} reports the results of applying ASAug advancements to Allspark~\cite{wang2024allspark} and CorrMatch~\cite{sun2024corrmatch}, along with comparisons with other leading methods in two dimension settings ($321$ and $513$). 
The inclusion of ASAug consistently enhances the performance of existing techniques by a large margin. Besides, all methods show improved results with a larger crop size. 
With the support of ASAug, Allspark experiences an increase of $mIoU$ of 1.2\% with a crop size of $513 \times 513$ and an increase of 4.16\% with a crop size of $321 \times 321$ when trained on 92 labeled samples. 
Additionally, it can be seen that ASAug performs effectively in various scenarios, ranging from 1/16 partially labeled to fully labeled cases. 
This highlights ASAug's effectiveness and establishes it as a versatile and valuable approach. 

% \newline
\textbf{Pascal VOC 2012 blender set.}
Table~\ref{tab:pascal_aug} shows the numerical analysis for the larger $blender$ set. Both Allspark~\cite{wang2024allspark} and CorrMatch~\cite{sun2024corrmatch} have demonstrated better performance with ASAug, which further verified the generalization of ASAug in all scenarios. At a 1/8 scale, Allspark's performance sees a 2.04\% improvement with ASAug, and CorrMatch results are similarly improved. An important finding is that our method significantly enhances performance, with results using a $321 \times 321$ crop size even outperforming some larger-scale $513 \times 513$ methods. Furthermore, we provide results within the same segmentation framework as U2PL \cite{wang2022semi}, including fully annotated labels. Our method achieves a new state-of-the-art performance using Allspark with ASAug. These results verified the effectiveness and generalization of our ASAug.

% 条形消融图
\begin{figure}[!t]
    % \vspace{-1.0em}
    \centering
    \includegraphics[width=1.0\linewidth]{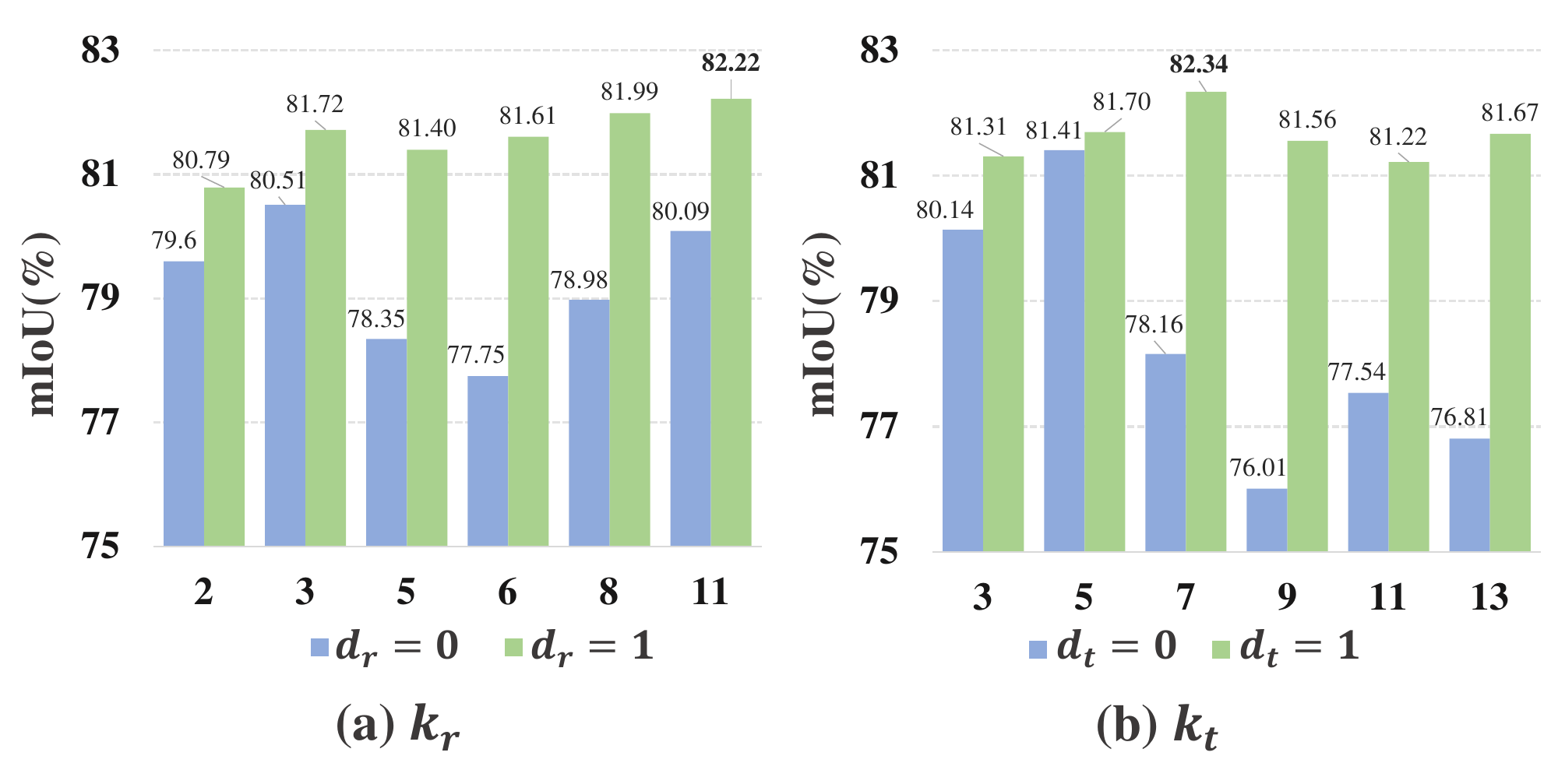}
    \vspace{-1.5em}
    \caption{Ablation study on EAW hyper-parameters $k_r$ and $k_t$ trained using the partitions of 1464, $d_t=d_r=1.0$ (Based on Allspark\cite{wang2024allspark}).}
    \label{fig_ablation_kd}
    \vspace{-1.0em}
\end{figure}

% \newline
\textbf{Cityscapes.}
Table~\ref{tab: city} evaluates the effectiveness of ASAug on the more challenging Cityscapes dataset, comparing it with other approaches. The evaluation uses a sliding window technique, showing that ASAug consistently delivers stable performance across different segments, even in complex urban scenes.
We can easily see that ASAug can readily outperform other methods, especially with scarce labels.  ASAug can improve Allspark by 4.16\%, 2.45\%, 2.85\%, 1.55\%, and 2.38\% under 1/16, 1/8, 1/4, 1/2, and full partition protocols, respectively. This impressive improvement further demonstrates the effectiveness and importance of our claim that spatial data augmentation is effective and should be adopted for SSSS.

\textbf{COCO.}
Table~\ref{tab: coco} shows the performance of ASAug on the COCO dataset, and the results demonstrate that ASAug outperforms Allspark in all partitioning cases (1/512, 1/256, 1/128, 1/64, and 1/32), and in particular, in the case of 1/32, 1/64, and 1/128 partitions respectively, it improves by 3.53\%, 2.75\% and 2.94\%, these experimental results further validate the effectiveness of spatial transformations on the multi-category datasets.

% \begin{table}[!t]
% \vspace{-0.5em}
% \footnotesize
% % \small
% % \normalsize
% \centering
% \caption{Ablation study on our ASAug. $\mathcal{A}_r$ and $\mathcal{A}_t$ represent the two main augmentation operations, respectively. Improvements to the {\color[HTML]{3531FF} supervised} baseline are highlighted in \red{red} (Based on DeepLabV3+~\cite{chen2017rethinking}).}
% \begin{tabular}{ccc|cc}
% % \hline
% \toprule[1pt]
% \rowcolor[HTML]{ECF4FF} 
% \multicolumn{3}{c|}{\cellcolor[HTML]{ECF4FF}ASAug} 
% & \multicolumn{2}{c}{\cellcolor[HTML]{ECF4FF}mIoU(\%)} \\ 
% % \hline
% \midrule
% MT& $\mathcal{A}_r$ & $\mathcal{A}_t$ & 732 & 366 
% \\
% \midrule
% % \hline
%    &    &    
%    & 66.26 {\color[HTML]{3531FF} (supervised)} 
%    & 60.13 {\color[HTML]{3531FF} (supervised)}
% \\
% \checkmark  &    &    
% & 69.54 {\color[HTML]{FE0000} ($\uparrow $3.28)}
% & 65.92 {\color[HTML]{FE0000} ($\uparrow $5.79)}
% \\
% \checkmark  & \checkmark  &    
% & 78.71 {\color[HTML]{FE0000} ($\uparrow $12.45)}     
% & 78.04 {\color[HTML]{FE0000} ($\uparrow $17.91)}      
% \\
% \checkmark  &    & \checkmark  
% & 78.18 {\color[HTML]{FE0000} ($\uparrow $11.92)}  
% & 77.30 {\color[HTML]{FE0000} ($\uparrow $17.17)}      
% \\ 
% \midrule
% % \hline
% \rowcolor[HTML]{ECF4FF} 
% \checkmark  & \checkmark  & \checkmark  & 
% \textbf{80.94 }
% {\color[HTML]{FE0000} ($\uparrow $14.68)} & 
% \textbf{79.98 }
% {\color[HTML]{FE0000} ($\uparrow $19.85)} \\ 
% \bottomrule[1pt]
% % \hline
% \end{tabular}
% \label{ablation_component}
% \vspace{-1.7em}
% \end{table}

% 消融实验
\subsection{Ablation Studies}
% \vspace{-0.5em}
\label{ablation}
% In this section, we perform ablation studies on the $classic$ Pascal VOC dataset at 732 (1/2) and 366 1464 labels.
We undertake multiple ablation studies to confirm the structure of the improvement approach introduced in ASAug. 
% For consistency across studies, we employ both DeepLabV3+~\cite{chen2017rethinking} and ResNet-101~\cite{he2016deep}, each pre-trained on ImageNet~\cite{deng2009imagenet} as the backbone frameworks. 
For consistency across studies, we used both CNN-based CorrMatch\cite{sun2024corrmatch} method and ViT-based Allspark\cite{wang2024allspark} method frameworks for ablation experiments.
We present our findings using the Pascal VOC 2012 $classic$ dataset, with a training dimension of 321 $\times $ 321.

% \
% \newline
\textbf{Effectiveness of ASAug.}
We begin by assessing the impact of individual components in ASAug, which are comprehensively presented in Table~\ref{ablation_component}. 
% The baseline results stem from the supervised method. 
The baseline results are from CorrMatch~ \cite{sun2024corrmatch}, and it is worth noting the point that the data presented in Table~\ref{ablation_component} are from our reproducible ablation experiments using CorrMatch.
% The term ``MT" denotes the traditional mean-teacher technique. 
Our streamlined augmentations, denoted as $\mathcal{A}_r$ and $\mathcal{A}_t$, significantly enhance performance. The full implementation, $\mathcal{T}_{ss}(\mathcal{A}_r,\mathcal{A}_t)$, exceeds the performance of either $\mathcal{A}_r$ or $\mathcal{A}_t$ when applied independently. This indicates that strong spatial augmentations $\mathcal{A}_r$ and $\mathcal{A}_t$ both serve effectively as regularizers, with their combination yielding even superior results.
% Fusing these two augmentations can enhance each component further and achieve peak performance.

\begin{figure}[!t]
    \vspace{-1.5em}
    \centering
    \includegraphics[width=1.0\linewidth]{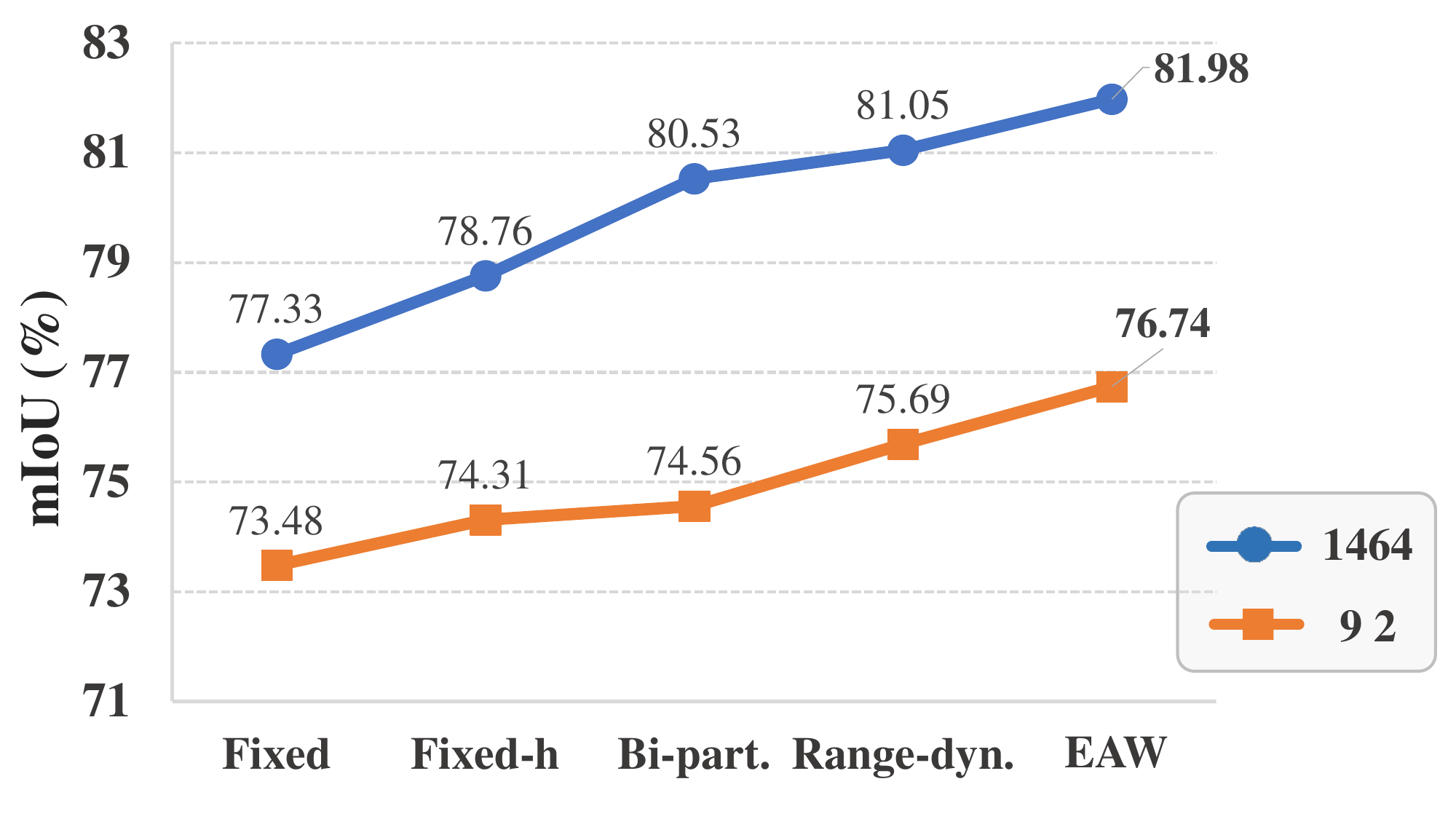}
    \vspace{-1.5em}
    \caption{Ablation study on mapping strategies. ``EAW" denotes our entropy-based adaptive dynamic mapping.}
    \label{fig_ablation_ms}
    \vspace{-1.0em}
\end{figure}

% \
% \newline
\textbf{Effectiveness of EAW.}
We verified the need to dynamically adjust the weight of the spatial increase according to the variance of the instance, and Fig.~\ref{fig_ablation_m} illustrates the effectiveness of EAW compared to other spatial transformations. (a) presents the results using a constant rotation angle for all samples, while (b) depicts those with a fixed translation ratio. A noticeable variance in model performance is observed at different angles, generally improving as the angle increases. Similarly, the augmentation effect varies across different translation ratios. In both scenarios, EAW's adaptive adjustment of augmentation distortion degree demonstrates superior results.

\textbf{Effects of hyper-parameters.}
We further analyze the performance of ASAug for $\mathcal{A}_r$ and $\mathcal{A}_t$ with various scaling parameters $k_r, k_t$ and validate the importance of the offset parameters $d_r, d_t$, as displayed in Fig.~\ref{fig_ablation_kd}. The findings demonstrate that incorporating $d_r$ and $d_t$ generally enhances model performance across different hyperparameter settings, notably in higher $k_r$ and medium $k_t$ scenarios (82.22\% and 82.34\% for $k_r = 11$ and $k_t = 7$, respectively). Fig.~\ref{fig_ablation_kd} also illustrates that no single parameter combination is consistently superior to others.
% To keep our approach simple and consistent, we set $[k_r, d_r]$ and $[k_t, d_t]$ by default to $[11, 1]$ and $[7, 1]$, respectively.

We assess the effect of the mapping method in Fig.~\ref{fig_ablation_ms}, focusing solely on rotation mapping $\mathcal{A}_r$. Here, ``EAW" stands for our entropy-driven adaptive weight. 
``Fixed" and ``Fixed-h" refer to spatial rotation transformations applied to images for a standard fixed-angle view (30°) and a higher fixed angle (80°), respectively.
``Bi-part." signifies a two-part mapping (chosen between Fixed or Fixed-h based on the entropy value), while ``Range-dyn." involves splitting the angular range into three parts (low, medium, and high) based on experimental results, then conducting dynamic partition mapping based on entropy values. 
Fig.~\ref{fig_ablation_ms} shows that rotation angles significantly affect the effectiveness of ASAug. Our EAW strategy surpasses the other typical mappings, highlighting its notable robustness in handling complex features and limited samples.

\textbf{Efficiency.}
The ASAug serves as a plug-and-play data augmentation tool that improves feature learning without requiring modifications to the existing training framework. Although it marginally extends the duration of the training, it significantly enhances performance and leaves the inference process unchanged. As demonstrated in the table below for the Pascal blender dataset (both involving 1085 iterations), our approach effectively balances enhanced performance with practical efficiency. (Indicator: $mIoU$)

\begin{table}[t]
\centering
\footnotesize
% \scriptsize
\vspace{-1.0em}
% \small
% \scriptsize
\caption{Ablation study of the efficiency. We give a comparison of running time and results before and after applying ASAug, based on CorrMatch\cite{sun2024corrmatch}. (Indicator: $mIoU$).}
\begin{tabular}{c|cccc}
% \hline
\toprule[1pt]
\rowcolor[HTML]{ECF4FF} 
& \textbf{\begin{tabular}[c]{@{}c@{}}
1 / 16\\(662)\end{tabular}} 
& \textbf{\begin{tabular}[c]{@{}c@{}}
1 / 8\\(1323)\end{tabular}} 
& \textbf{\begin{tabular}[c]{@{}c@{}}
1 / 4\\(2646)\end{tabular}} 
& \textbf{\begin{tabular}[c]{@{}c@{}}
1 / 2\\(5291)\end{tabular}} 
\\ 
\midrule
% \hline
CorrMatch                     
& 32min25s                                             
& 30min06s                                              
& 25min48s                                             
& 17min25s         \\ 
% \midrule
% \hline
\rowcolor[HTML]{ECF4FF} 
\textbf{\begin{tabular}[c]{@{}c@{}}
% CorrMatch \\
w/ ASAug\end{tabular} }
& 39min02s                                              
& 36min02s                                              
& 30min31s                                             
& 21min05s  \\ 
% \hline
\midrule
% \rowcolor[HTML]{ECF4FF} 
\textbf{Time diff.}     
& \textbf{+ 6min37s }           
& \textbf{+ 5min56s}     
& \textbf{+ 4min43s}          
& \textbf{+ 3min40s}             \\ 
% \hline
\midrule
% \rowcolor[HTML]{ECF4FF} 
% \textbf{mIoU(\%) increase}
\textbf{
\begin{tabular}[c]{@{}c@{}} 
{\color[HTML]{3531FF} \textbf{$\uparrow $ $\bigtriangleup$ (\%) }}
\end{tabular}}
& \textbf{\color[HTML]{3531FF} $\uparrow $ 1.11 } 
& \textbf{\color[HTML]{3531FF} $\uparrow $ 1.13}     
& \textbf{\color[HTML]{3531FF} $\uparrow $ 0.77}   
& \textbf{\color[HTML]{3531FF} $\uparrow $ 0.55} \\
% \hline
\bottomrule[1pt]
\end{tabular}
\vspace{-1.0em}
\end{table}

\begin{figure}[!b]
    \vspace{-1.0em}
    \centering
    \includegraphics[width=1.0\linewidth,height=5cm]{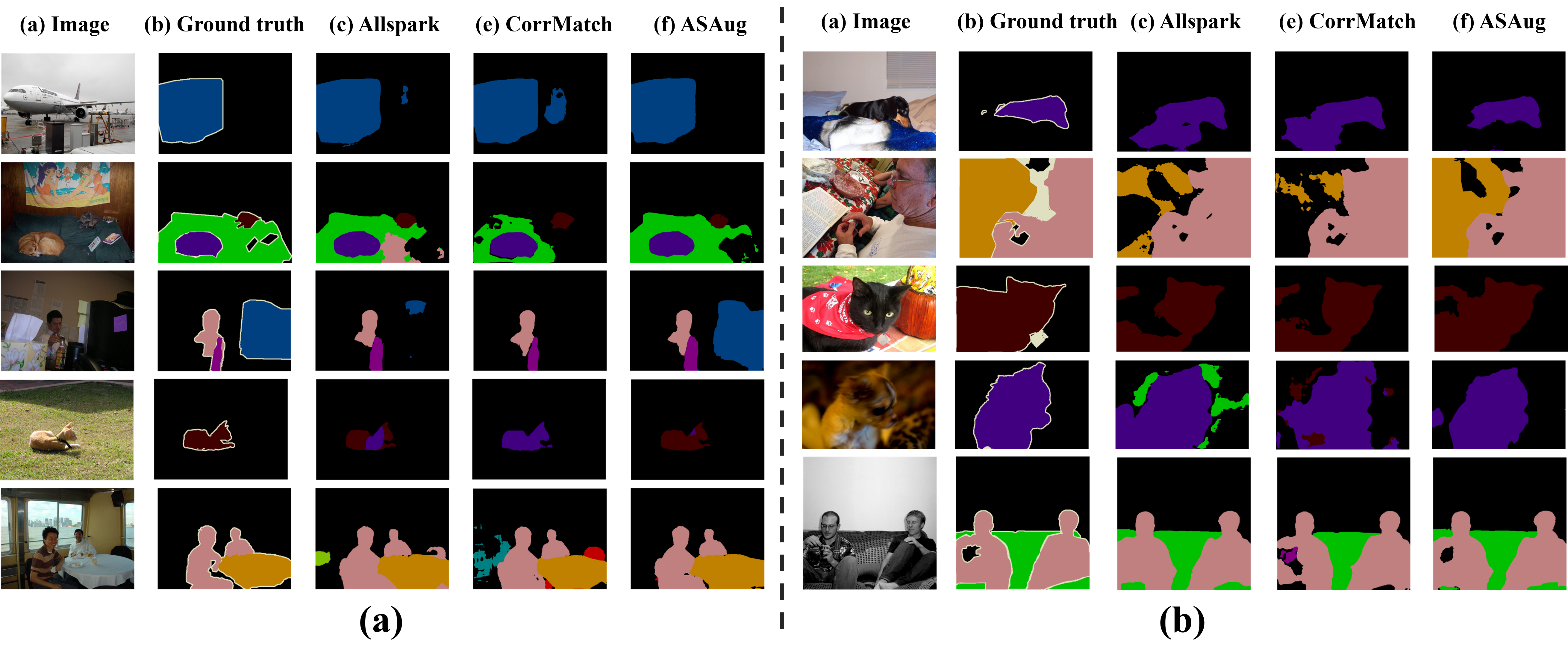}
     \vspace{-1.7em}
    \caption{Visualization of the segmentation results on Pascal validation set, compared with Allspark~\cite{wang2024allspark} and CorrMatch~\cite{sun2024corrmatch}.}
    \label{fig_res_pascal}
    % \includegraphics[width=0.8\linewidth, height=5cm]{fig/struct.png}
    % \vspace{-1.0em}
\end{figure}

\subsection{Qualitative Analysis}\label{QA}
% \vspace{-0.5em}
% \
% \newline
% \noindent
\textbf{Visualization on the Pascal dataset.}
\label{sec:rationale}
We visually compare the ASAug results with Allspark\cite{wang2024allspark} and CorrMatch\cite{sun2024corrmatch} for segmenting the Pascal validation datasets in Fig.~\ref{fig_res_pascal}. 
The ASAug technique shows a higher level of segmentation accuracy, regularly surpassing the performances of both Allspark and CorrMatch across various complex conditions. It offers enhanced capabilities in identifying object classes and boundaries, preserving detailed features, and reducing incorrect background predictions. For instance, the outlines of aircraft ((a)-Row 1), animals ((a)-Rows 2/4 and (b)-Rows 1/3/4), and people ((a)-Rows 3/5 and (b)-Row 5) are more precise and align closely with the actual situation. 
Additionally, ASAug reduces unnecessary background inclusions in the output, noticeable with the cat in the grass ((a)-Row 4) and the sleeping dog ((a)-Row 5). 
In scenes with multiple categories ((a)-Rows 2/5 and (b) Row 7), ASAug provides a more accurate segmentation of individual characters, whereas other methods generate some degree of noisy predictions, offering clearer outputs. 
These findings underscore the efficacy of ASAug's spatial improvements and consistency-driven regularization strategy, which enhance ASAug's performance across straightforward and challenging segmentation tasks.

\begin{figure}[!t]
    % \vspace{-2.0em}
    \centering
    \includegraphics[width=1\linewidth,height=5cm]{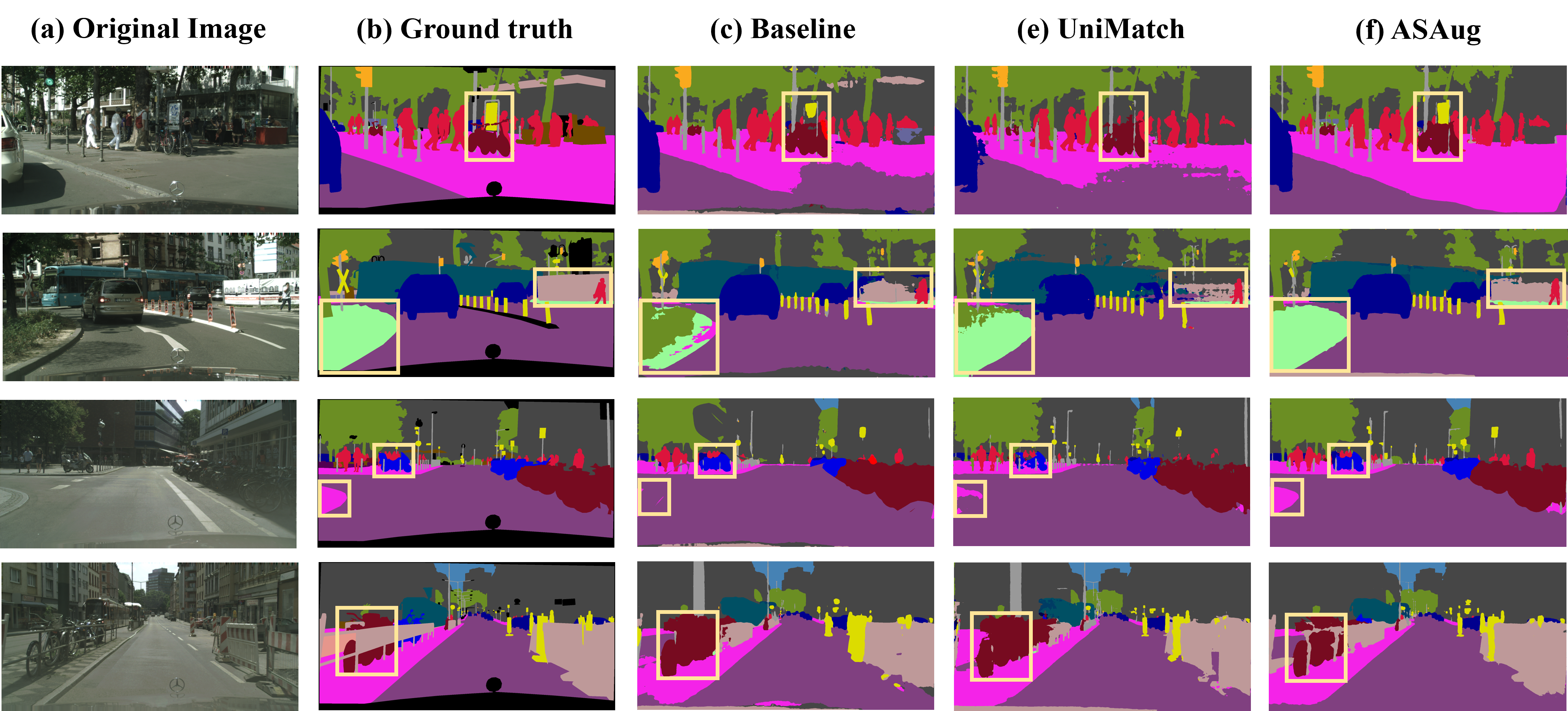}
    % \vspace{-0.8em}
    \caption{Visualization of sample predictions from Cityscapes, highlighting the improved segmentation accuracy achieved with ASAug.}
    \label{fig_res_city}
    \vspace{-1.0em}
\end{figure}

% \noindent
\textbf{Visualization on the Cityscapes dataset.}
Fig.~\ref{fig_res_city} presents visualization results on the Cityscapes dataset trained with 744 labeled samples, to illustrate the enhanced performance of ASAug. As our baseline, we utilized the original self-training model, which was trained solely with labeled data. 
The baseline results in some categories having vague edges, leading to lost details, such as in the boundary of lawns, trees, and buildings, with evident misclassifications and reduced detection accuracy. 
While UniMatch~\cite{yang2023revisiting} somewhat enhances edge quality, it still struggles with misclassification in intricate regions like those with pedestrians and road objects. 
In contrast, ASAug's outcomes are more aligned with the ground truth, particularly achieving higher segmentation accuracy for small objects like pedestrians and vehicles, showcasing its superior capability in detailing regions. 
This advancement may be attributed to ASAug's unique focus on positional information, which enables it to adapt to different scene variations, thereby improving detail discrimination while maintaining overall consistency, allowing ASAug to outperform other methods in terms of edge preservation and complex regions, validating its effectiveness in improving model generalization ability and detail fidelity.
Nonetheless, segmentation in complex category stacking remains challenging. We anticipate future improvements to ASAug for such scenarios.

% 结果图
% 图片的标注大小

% 参数消融--表格
% \begin{table}[!t]
% \centering
% % \small
% \footnotesize
% \caption{Ablation study on mapping hyper-parameters $k_r$ and $k_t$ trained using 1464 labeled data. $d_t$ and $d_r$ are defaulted to 1.0.}
% \begin{tabular}{c|cccccc}
% % \hline
% \toprule[1pt]
% \rowcolor[HTML]{ECF4FF} 
% % \textbf{$\mathcal{A}_r$} 
% \textbf{$k_r$} 
% & 2   & 3   & 5   & 6   &  8  & 11    \\ 
% \midrule
% % \hline
% w/o $d_r$      
% & 79.60 & 80.51 & 78.35   & 77.75   & 78.98   & 80.09  \\
% w/ $d_r$        
% & 80.79   & 81.72   & 81.40  & 81.61   & 81.99 
% &\textbf{82.22} \\ 
% \midrule[0.75pt]
% \midrule[0.75pt]
% % \hline
% \rowcolor[HTML]{ECF4FF} 
% % \textbf{$\mathcal{A}_t$} 
% \textbf{$k_t$} 
% & 3 & 5 & 7  & 9  & 11 & 13    \\ 
% \midrule
% % \hline
% w/o $d_t$  
% & 80.14 & 81.41  & 78.16  & 76.01  & 77.54   & 76.81       \\
% w/ $d_t$       
% & 81.31  & 81.70  & \textbf{82.34} & 81.56  & 81.22  & 81.67 \\ 
% \bottomrule[1pt]
% % \hline
% \end{tabular}
% \label{ablation_parameters}
% \end{table}

\section{Conclusion}
In this paper, we present ASAug, a simple yet effective method designed to address the challenges of SSSS. Unlike recent approaches that primarily rely on intensity-based augmentations, our findings demonstrate that strong spatial augmentations can significantly improve SSSS performance. ASAug utilizes translation and rotation as data augmentation techniques, promoting better generalization in the presence of dynamic disturbances. To ensure stability, we incorporate an entropy-based adaptive weight strategy to handle more challenging cases. ASAug achieves substantial improvements over state-of-the-art, without the need for complex architectures. 

We believe that shifting the focus from intensity-based augmentations to spatial techniques, such as translation and rotation, will lay a strong foundation for future SSSS research. Additionally, combining strong spatial augmentation with other augmentation strategies may show great potential.

%参考文献及格式
{
    % \small
    \footnotesize
    \bibliographystyle{IEEEtran} 
    \bibliography{myBib}

\begin{thebibliography}{10}
\providecommand{\url}[1]{#1}
\csname url@samestyle\endcsname
\providecommand{\newblock}{\relax}
\providecommand{\bibinfo}[2]{#2}
\providecommand{\BIBentrySTDinterwordspacing}{\spaceskip=0pt\relax}
\providecommand{\BIBentryALTinterwordstretchfactor}{4}
\providecommand{\BIBentryALTinterwordspacing}{\spaceskip=\fontdimen2\font plus
\BIBentryALTinterwordstretchfactor\fontdimen3\font minus \fontdimen4\font\relax}
\providecommand{\BIBforeignlanguage}[2]{{%
\expandafter\ifx\csname l@#1\endcsname\relax
\typeout{** WARNING: IEEEtran.bst: No hyphenation pattern has been}%
\typeout{** loaded for the language `#1'. Using the pattern for}%
\typeout{** the default language instead.}%
\else
\language=\csname l@#1\endcsname
\fi
#2}}
\providecommand{\BIBdecl}{\relax}
\BIBdecl

\bibitem{ma2025dual}
Q.~Ma, Z.~Zhang, P.~Qiao, Y.~Wang, R.~Ji, C.~Liu, and J.~Chen, ``Dual-level masked semantic inference for semi-supervised semantic segmentation,'' \emph{TMM}, 2025.

\bibitem{zhang2025frequency}
S.~Zhang, D.~Kong, Y.~Xing, Y.~Lu, L.~Ran, G.~Liang, H.~Wang, and Y.~Zhang, ``Frequency-guided spatial adaptation for camouflaged object detection,'' \emph{TMM}, vol.~27, pp. 72--83, 2025.

\bibitem{ding2024clustering}
Y.~Ding, L.~Li, W.~Wang, and Y.~Yang, ``Clustering propagation for universal medical image segmentation,'' in \emph{CVPR}, 2024.

\bibitem{lran2024DDF}
L.~Ran, L.~Wang, T.~Zhuo, Y.~Xing, and Y.~Zhang, ``{DDF}: A novel dual-domain image fusion strategy for remote sensing image semantic segmentation with unsupervised domain adaptation,'' \emph{TGRS}, 2024.

\bibitem{guo2024vanishing}
D.~Guo, D.-P. Fan, T.~Lu, C.~Sakaridis, and L.~Van~Gool, ``Vanishing-point-guided video semantic segmentation of driving scenes,'' in \emph{CVPR}, 2024.

\bibitem{chen2025towards}
B.~Chen, Z.~Ye, Y.~Liu, X.~Fang, G.~Lu, S.~Xie, and X.~Li, ``Towards robust semi-supervised distribution alignment against label distribution shift with noisy annotations,'' \emph{TMM}, 2025.

\bibitem{lu2025uncertainty}
X.~Lu, L.~Li, L.~Jiao, X.~Liu, F.~Liu, W.~Ma, and S.~Yang, ``Uncertainty-aware semi-supervised learning segmentation for remote sensing images,'' \emph{TMM}, 2025.

\bibitem{pseudolabel}
L.~Ran, Y.~Li, G.~Liang, and Y.~Zhang, ``Pseudo labeling methods for semi-supervised semantic segmentation: A review and future perspectives,'' \emph{TCSVT}, vol.~35, no.~4, pp. 3054--3080, 2025.

\bibitem{hu2024multi}
K.~Hu, X.~Chen, Z.~Chen, Y.~Zhang, and X.~Gao, ``Multi-perspective pseudo-label generation and confidence-weighted training for semi-supervised semantic segmentation,'' \emph{TMM}, 2024.

\bibitem{xiao2024multi}
H.~Xiao, Y.~Hong, L.~Dong, D.~Yan, J.~Xiong, J.~Zhuang, D.~Liang, and C.~Peng, ``Multi-level label correction by distilling proximate patterns for semi-supervised semantic segmentation,'' \emph{TMM}, 2024.

\bibitem{zou2021PseudoSeg}
Y.~Zou, Z.~Zhang, H.~Zhang, C.-L. Li, X.~Bian, J.-B. Huang, and T.~Pfister, ``Pseudoseg: Designing pseudo labels for semantic segmentation,'' in \emph{ICLR}, 2021.

\bibitem{lee2021anti}
J.~Lee, E.~Kim, and S.~Yoon, ``Anti-adversarially manipulated attributions for weakly and semi-supervised semantic segmentation,'' in \emph{CVPR}, 2021.

\bibitem{yang2023revisiting}
L.~Yang, L.~Qi, L.~Feng, W.~Zhang, and Y.~Shi, ``Revisiting weak-to-strong consistency in semi-supervised semantic segmentation,'' in \emph{CVPR}, 2023.

\bibitem{yuan2021simple}
J.~Yuan, Y.~Liu, C.~Shen, Z.~Wang, and H.~Li, ``A simple baseline for semi-supervised semantic segmentation with strong data augmentation,'' in \emph{ICCV}, 2021.

\bibitem{zhao2023augmentation}
Z.~Zhao, L.~Yang, S.~Long, J.~Pi, L.~Zhou, and J.~Wang, ``Augmentation matters: {A} simple-yet-effective approach to semi-supervised semantic segmentation,'' in \emph{CVPR}, 2023.

\bibitem{Cubuk_2019_CVPR}
E.~D. Cubuk, B.~Zoph, D.~Mane, V.~Vasudevan, and Q.~V. Le, ``{AutoAugment}: {Learning} augmentation strategies from data,'' in \emph{CVPR}, Jun. 2019.

\bibitem{cubuk2020randaugment}
E.~D. Cubuk, B.~Zoph, J.~Shlens, and Q.~V. Le, ``Randaugment: {Practical} automated data augmentation with a reduced search space,'' in \emph{CVPRW}, 2020.

\bibitem{muller2021trivialaugment}
S.~G. Müller and F.~Hutter, ``Trivialaugment: {Tuning}-free yet state-of-the-art data augmentation,'' in \emph{ICCV}, 2021.

\bibitem{ouali2020semi}
Y.~Ouali, C.~Hudelot, and M.~Tami, ``Semi-supervised semantic segmentation with cross-consistency training,'' in \emph{CVPR}, 2020.

\bibitem{zhou2021c3}
Y.~Zhou, H.~Xu, W.~Zhang, B.~Gao, and P.-A. Heng, ``C3-semiseg: Contrastive semi-supervised segmentation via cross-set learning and dynamic class-balancing,'' in \emph{ICCV}, 2021.

\bibitem{feng2022dmt}
Z.~Feng, Q.~Zhou, Q.~Gu, X.~Tan, G.~Cheng, X.~Lu, J.~Shi, and L.~Ma, ``Dmt: Dynamic mutual training for semi-supervised learning,'' \emph{PR}, 2022.

\bibitem{liu2021bootstrapping}
S.~Liu, S.~Zhi, E.~Johns, and A.~J. Davison, ``Bootstrapping semantic segmentation with regional contrast,'' \emph{ICLR}, 2022.

\bibitem{chen2021semi}
X.~Chen, Y.~Yuan, G.~Zeng, and J.~Wang, ``Semi-supervised semantic segmentation with cross pseudo supervision,'' in \emph{CVPR}, 2021.

\bibitem{yang2022st++}
L.~Yang, W.~Zhuo, L.~Qi, Y.~Shi, and Y.~Gao, ``St++: Make self-training work better for semi-supervised semantic segmentation,'' in \emph{CVPR}, 2022.

\bibitem{kwon2022semi}
D.~Kwon and S.~Kwak, ``Semi-supervised semantic segmentation with error localization network,'' in \emph{CVPR}, 2022.

\bibitem{liu2022perturbed}
Y.~Liu, Y.~Tian, Y.~Chen, F.~Liu, V.~Belagiannis, and G.~Carneiro, ``Perturbed and strict mean teachers for semi-supervised semantic segmentation,'' in \emph{CVPR}, 2022.

\bibitem{wang2022semi}
Y.~Wang, H.~Wang, Y.~Shen, J.~Fei, W.~Li, G.~Jin, L.~Wu, R.~Zhao, and X.~Le, ``Semi-supervised semantic segmentation using unreliable pseudo-labels,'' in \emph{CVPR}, 2022.

\bibitem{fan2022ucc}
J.~Fan, B.~Gao, H.~Jin, and L.~Jiang, ``Ucc: Uncertainty guided cross-head co-training for semi-supervised semantic segmentation,'' in \emph{CVPR}, 2022.

\bibitem{wang2023conflict}
Z.~Wang, Z.~Zhao, X.~Xing, D.~Xu, X.~Kong, and L.~Zhou, ``Conflict-based cross-view consistency for semi-supervised semantic segmentation,'' in \emph{CVPR}, 2023.

\bibitem{zhao2023instance}
Z.~Zhao, S.~Long, J.~Pi, J.~Wang, and L.~Zhou, ``Instance-specific and model-adaptive supervision for semi-supervised semantic segmentation,'' in \emph{CVPR}, 2023.

\bibitem{wu2023querying}
L.~Wu, L.~Fang, X.~He, M.~He, J.~Ma, and Z.~Zhong, ``Querying labeled for unlabeled: Cross-image semantic consistency guided semi-supervised semantic segmentation,'' \emph{TPAMI}, 2023.

\bibitem{sun2024corrmatch}
B.~Sun, Y.~Yang, L.~Zhang, M.-M. Cheng, and Q.~Hou, ``Corrmatch: Label propagation via correlation matching for semi-supervised semantic segmentation,'' in \emph{CVPR}, 2024.

\bibitem{wang2024allspark}
H.~Wang, Q.~Zhang, Y.~Li, and X.~Li, ``Allspark: Reborn labeled features from unlabeled in transformer for semi-supervised semantic segmentation,'' in \emph{CVPR}, 2024.

\bibitem{everingham2010pascal}
M.~Everingham, L.~Van~Gool, C.~K. Williams, J.~Winn, and A.~Zisserman, ``The pascal visual object classes (voc) challenge,'' \emph{IJCV}, 2010.

\bibitem{cordts2016cityscapes}
M.~Cordts, M.~Omran, S.~Ramos, T.~Rehfeld, M.~Enzweiler, R.~Benenson, U.~Franke, S.~Roth, and B.~Schiele, ``The cityscapes dataset for semantic urban scene understanding,'' in \emph{CVPR}, 2016.

\bibitem{lin2014microsoft}
T.-Y. Lin, M.~Maire, S.~Belongie, J.~Hays, P.~Perona, D.~Ramanan, P.~Doll{\'a}r, and C.~L. Zitnick, ``Microsoft coco: Common objects in context,'' in \emph{ECCV}.\hskip 1em plus 0.5em minus 0.4em\relax Springer, 2014, pp. 740--755.

\bibitem{xu2021dash}
Y.~Xu, L.~Shang, J.~Ye, Q.~Qian, Y.-F. Li, B.~Sun, H.~Li, and R.~Jin, ``Dash: Semi-supervised learning with dynamic thresholding,'' in \emph{ICLR}, 2021.

\bibitem{lee2013pseudo}
D.-H. Lee \emph{et~al.}, ``Pseudo-label: The simple and efficient semi-supervised learning method for deep neural networks,'' in \emph{Workshop on challenges in representation learning, ICML}, 2013.

\bibitem{adasemicd}
L.~Ran, D.~Wen, T.~Zhuo, S.~Zhang, X.~Zhang, and Y.~Zhang, ``Adasemicd: An adaptive semi-supervised change detection method based on pseudo-label evaluation,'' \emph{TGRS}, vol.~63, pp. 1--14, 2025.

\bibitem{sohn2020fixmatch}
K.~Sohn, D.~Berthelot, N.~Carlini, Z.~Zhang, H.~Zhang, C.~A. Raffel, E.~D. Cubuk, A.~Kurakin, and C.-L. Li, ``Fixmatch: Simplifying semi-supervised learning with consistency and confidence,'' \emph{NeurIPS}, 2020.

\bibitem{NEURIPS2021_995693c1}
B.~Zhang, Y.~Wang, W.~Hou, H.~Wu, J.~Wang, M.~Okumura, and T.~Shinozaki, ``Flexmatch: Boosting semi-supervised learning with curriculum pseudo labeling,'' in \emph{NeurIPS}, 2021.

\bibitem{pelaez2023survey}
A.~Pel{\'a}ez-Vegas, P.~Mesejo, and J.~Luengo, ``A survey on semi-supervised semantic segmentation,'' \emph{arXiv:2302.09899}, 2023.

\bibitem{bachman2014learning}
P.~Bachman, O.~Alsharif, and D.~Precup, ``Learning with pseudo-ensembles,'' \emph{NeurIPS}, 2014.

\bibitem{oord2018representation}
A.~v.~d. Oord, Y.~Li, and O.~Vinyals, ``Representation learning with contrastive predictive coding,'' \emph{arXiv:1807.03748}, 2018.

\bibitem{goodfellow2014generative}
I.~Goodfellow, J.~Pouget-Abadie, M.~Mirza, B.~Xu, D.~Warde-Farley, S.~Ozair, A.~Courville, and Y.~Bengio, ``Generative adversarial nets,'' \emph{NeurIPS}, 2014.

\bibitem{hou2022semi}
J.~Hou, X.~Ding, and J.~D. Deng, ``Semi-supervised semantic segmentation of vessel images using leaking perturbations,'' in \emph{WACV}, 2022.

\bibitem{xu2022self}
Y.~Xu, F.~He, B.~Du, D.~Tao, and L.~Zhang, ``Self-ensembling gan for cross-domain semantic segmentation,'' \emph{TMM}, vol.~25, pp. 7837--7850, 2022.

\bibitem{souly2017semi}
N.~Souly, C.~Spampinato, and M.~Shah, ``Semi supervised semantic segmentation using generative adversarial network,'' in \emph{ICCV}, 2017.

\bibitem{li2021semantic}
D.~Li, J.~Yang, K.~Kreis, A.~Torralba, and S.~Fidler, ``Semantic segmentation with generative models: Semi-supervised learning and strong out-of-domain generalization,'' in \emph{CVPR}, 2021.

\bibitem{cao2022adversarial}
C.~Cao, T.~Lin, D.~He, F.~Li, H.~Yue, J.~Yang, and E.~Ding, ``Adversarial dual-student with differentiable spatial warping for semi-supervised semantic segmentation,'' \emph{TCSVT}, 2022.

\bibitem{jin2021adversarial}
G.~Jin, C.~Liu, and X.~Chen, ``Adversarial network integrating dual attention and sparse representation for semi-supervised semantic segmentation,'' \emph{Inf. Process. Manag.}, 2021.

\bibitem{xie2023boosting}
H.~Xie, C.~Wang, M.~Zheng, M.~Dong, S.~You, C.~Fu, and C.~Xu, ``Boosting semi-supervised semantic segmentation with probabilistic representations,'' in \emph{AAAI}, 2023.

\bibitem{zoupseudoseg}
Y.~Zou, Z.~Zhang, H.~Zhang, C.-L. Li, X.~Bian, J.-B. Huang, and T.~Pfister, ``Pseudoseg: Designing pseudo labels for semantic segmentation,'' in \emph{ICLR}, 2021.

\bibitem{wang2023hunting}
X.~Wang, B.~Zhang, L.~Yu, and J.~Xiao, ``Hunting sparsity: Density-guided contrastive learning for semi-supervised semantic segmentation,'' in \emph{CVPR}, 2023.

\bibitem{zhong2021pixel}
Y.~Zhong, B.~Yuan, H.~Wu, Z.~Yuan, J.~Peng, and Y.-X. Wang, ``Pixel contrastive-consistent semi-supervised semantic segmentation,'' in \emph{ICCV}, 2021.

\bibitem{ouali2020overview}
Y.~Ouali, C.~Hudelot, and M.~Tami, ``An overview of deep semi-supervised learning,'' \emph{arXiv:2006.05278}, 2020.

\bibitem{yun2019cutmix}
S.~Yun, D.~Han, S.~J. Oh, S.~Chun, J.~Choe, and Y.~Yoo, ``Cutmix: Regularization strategy to train strong classifiers with localizable features,'' in \emph{ICCV}, 2019.

\bibitem{olsson2021classmix}
V.~Olsson, W.~Tranheden, J.~Pinto, and L.~Svensson, ``Classmix: Segmentation-based data augmentation for semi-supervised learning,'' in \emph{WACV}, 2021.

\bibitem{chen2021complexmix}
Y.~Chen, X.~Ouyang, K.~Zhu, and G.~Agam, ``Complexmix: Semi-supervised semantic segmentation via mask-based data augmentation,'' in \emph{ICIP}, 2021.

\bibitem{grubivsic2023revisiting}
I.~Grubi{\v{s}}i{\'c}, M.~Or{\v{s}}i{\'c}, and S.~{\v{S}}egvi{\'c}, ``Revisiting consistency for semi-supervised semantic segmentation,'' \emph{Sensors}, 2023.

\bibitem{grubivsic2021baseline}
------, ``A baseline for semi-supervised learning of efficient semantic segmentation models,'' in \emph{MVA}, 2021.

\bibitem{tarvainen2017mean}
A.~Tarvainen and H.~Valpola, ``Mean teachers are better role models: Weight-averaged consistency targets improve semi-supervised deep learning results,'' \emph{NeurIPS}, 2017.

\bibitem{lai2021semi}
X.~Lai, Z.~Tian, L.~Jiang, S.~Liu, H.~Zhao, L.~Wang, and J.~Jia, ``Semi-supervised semantic segmentation with directional context-aware consistency,'' in \emph{CVPR}, 2021.

\bibitem{kong2023pruning}
H.~Kong, G.-H. Lee, S.~Kim, and S.-W. Lee, ``Pruning-guided curriculum learning for semi-supervised semantic segmentation,'' in \emph{WACV}, 2023.

\bibitem{french2019semi}
G.~French, S.~Laine, T.~Aila, M.~Mackiewicz, and G.~Finlayson, ``Semi-supervised semantic segmentation needs strong, varied perturbations,'' \emph{arXiv:1906.01916}, 2019.

\bibitem{jin2022semi}
Y.~Jin, J.~Wang, and D.~Lin, ``Semi-supervised semantic segmentation via gentle teaching assistant,'' \emph{NeurIPS}, 2022.

\bibitem{ma2023enhanced}
J.~Ma, C.~Wang, Y.~Liu, L.~Lin, and G.~Li, ``Enhanced soft label for semi-supervised semantic segmentation,'' in \emph{ICCV}, 2023.

\bibitem{everingham2015pascal}
M.~Everingham, S.~A. Eslami, L.~Van~Gool, C.~K. Williams, J.~Winn, and A.~Zisserman, ``The pascal visual object classes challenge: A retrospective,'' \emph{IJCV}, 2015.

\bibitem{hariharan2011semantic}
B.~Hariharan, P.~Arbel{\'a}ez, L.~Bourdev, S.~Maji, and J.~Malik, ``Semantic contours from inverse detectors,'' in \emph{ICCV}, 2011.

\bibitem{hu2021semi}
H.~Hu, F.~Wei, H.~Hu, Q.~Ye, J.~Cui, and L.~Wang, ``Semi-supervised semantic segmentation via adaptive equalization learning,'' \emph{NeurIPS}, 2021.

\bibitem{li2023cfcg}
S.~Li, Y.~He, W.~Zhang, W.~Zhang, X.~Tan, J.~Han, E.~Ding, and J.~Wang, ``Cfcg: Semi-supervised semantic segmentation via cross-fusion and contour guidance supervision,'' in \emph{ICCV}, 2023.

\bibitem{yuan2023semi}
J.~Yuan, J.~Ge, Z.~Wang, and Y.~Liu, ``Semi-supervised semantic segmentation with mutual knowledge distillation,'' in \emph{ACM MM}, 2023.

\bibitem{liang2023logic}
C.~Liang, W.~Wang, J.~Miao, and Y.~Yang, ``Logic-induced diagnostic reasoning for semi-supervised semantic segmentation,'' in \emph{ICCV}, 2023, pp. 16\,197--16\,208.

\bibitem{hu2024training}
X.~Hu, L.~Jiang, and B.~Schiele, ``Training vision transformers for semi-supervised semantic segmentation,'' in \emph{CVPR}, 2024.

\bibitem{dong2024boundary}
J.~Dong, Z.~Meng, D.~Liu, J.~Liu, Z.~Zhao, and F.~Su, ``Boundary-refined prototype generation: A general end-to-end paradigm for semi-supervised semantic segmentation,'' \emph{EAAI}, 2024.

\bibitem{chen2017rethinking}
L.-C. Chen, G.~Papandreou, F.~Schroff, and H.~Adam, ``Rethinking atrous convolution for semantic image segmentation,'' \emph{arXiv:1706.05587}, 2017.

\bibitem{he2016deep}
K.~He, X.~Zhang, S.~Ren, and J.~Sun, ``Deep residual learning for image recognition,'' in \emph{CVPR}, 2016.

\bibitem{deng2009imagenet}
J.~Deng, W.~Dong, R.~Socher, L.-J. Li, K.~Li, and L.~Fei-Fei, ``Imagenet: A large-scale hierarchical image database,'' in \emph{CVPR}, 2009.

\bibitem{xie2021segformer}
E.~Xie, W.~Wang, Z.~Yu, A.~Anandkumar, J.~M. Alvarez, and P.~Luo, ``Segformer: Simple and efficient design for semantic segmentation with transformers,'' \emph{NeurIPS}, 2021.

\end{thebibliography}
    
}

\end{document}